\documentclass{article}
\pdfoutput=1

\PassOptionsToPackage{numbers, compress}{natbib}

\usepackage[preprint]{neurips_2023}

\usepackage{amsmath,amssymb,amsfonts}
\usepackage{algorithmic}
\usepackage{textcomp}
\usepackage{svg}
\usepackage{cite}
\usepackage[utf8]{inputenc} 
\usepackage[T1]{fontenc}    
\usepackage{hyperref}       
\usepackage{url}            
\usepackage{amssymb}
\usepackage{latexsym}
\usepackage{graphicx}
\usepackage{amsmath}
\usepackage{subfigure}
\usepackage{booktabs}       
\usepackage{amsfonts}       
\usepackage{nicefrac}       
\usepackage{microtype}      
\usepackage{xcolor}         
\usepackage{framed,multirow}

\hypersetup{
colorlinks=true,
linkcolor=black
}

\usepackage{listings}

\usepackage{makecell} 
\usepackage{pifont} 
\usepackage{enumerate}
\usepackage{enumitem}
\usepackage{comment}
\usepackage{ragged2e}
\usepackage{tikz}
\newcommand*\circled[1]{\tikz[baseline=(char.base)]{
            \node[shape=circle,draw,inner sep=1pt] (char) {#1};}}

\hypersetup{
    colorlinks=true,
    linkcolor=red,
    filecolor=green,      
    citecolor=green,
}

\title{PnPNet: Pull-and-Push Networks for Volumetric Segmentation with Boundary Confusion}

\author{
        Xin You$^{1}$\thanks{First Author}, Ming Ding$^{2}$, Minghui Zhang$^{1}$, Hanxiao Zhang$^{1}$, Yi Yu$^{2}$$^{\dag}$, Jie Yang$^{1}$$^{\dag}$, Yun Gu$^{1}$\thanks{Corresponding Author}\\
  $^{1}$Institute of Medical Robotics, Shanghai Jiao Tong University, Shanghai, China \\
$^{2}$Shanghai Xinhua Hospital, Shanghai Jiao Tong University, Shanghai, China \\
    \texttt{\{sjtu\_youxin, jieyang, geron762\}@sjtu.edu.cn} \\
}

\begin{document}

\maketitle

\begin{abstract}
Precise boundary segmentation of volumetric images is a critical task for image-guided diagnosis and computer-assisted intervention, especially for boundary confusion in clinical practice. However, U-shape deep networks cannot effectively resolve this challenge due to the lack of boundary shape constraints. Besides, existing methods of refining boundaries overemphasize the slender structure, which results in the overfitting phenomenon due to neural networks' limited abilities to model tiny objects. In this paper, we reconceptualize the mechanism of boundary generation by encompassing the dynamics of interactions with adjacent regions. Moreover, we propose a unified network termed PnPNet to model shape characteristics of the confused boundary region. The core ingredients of PnPNet contain the pushing and pulling branches. Specifically, based on diffusion theory, we devise the semantic difference guidance module (SDM) from the pushing branch to squeeze the boundary region. Explicit and implicit differential information inside SDM will significantly boost the representation abilities for inter-class boundaries. Additionally, motivated by the K-means algorithm, the class clustering module (CCM) from the pulling branch is introduced to stretch the intersected boundary region. Thus, pushing and pulling branches will shrink and enlarge the boundary uncertainty respectively. They furnish two adversarial forces to enhance models' representation capabilities for the boundary region, then promote models to output a more precise delineation of inter-class boundaries. We carry out quantitative and qualitative experiments on three challenging public datasets and one in-house dataset, containing three different types of boundary confusion in model predictions. Experimental results demonstrate the superiority of PnPNet over other segmentation networks, especially on the evaluation metrics of Hausdorff Distance and Average Symmetric Surface Distance. Besides, pushing and pulling branches can serve as plug-and-play modules to enhance classic U-shape baseline models. Source codes are available at \url{https://github.com/AlexYouXin/PnPNet}.
\end{abstract}

\section{Introduction}
\label{sec1}
Volumetric medical image segmentation plays an essential role in numerous clinical applications \citep{isensee2021nnu, aerts2014decoding}, including artificial intelligence in diagnostic support systems \citep{bernard2018deep, de2018clinically}, therapy planning support \citep{nikolov2018deep}, intra-operative assistance \citep{hollon2020near} and tumor growth monitoring \citep{kickingereder2019automated} etc. The process of segmentation is defined as the precise pixel-wise annotation for regions of interest (ROIs) \citep{azad2022medical}. Further, the essence of precise ROI segmentation refers to the accurate localization of foreground boundaries \citep{pham2000current}, which is the fundamental challenge of medical image segmentation.

Recently, U-shape neural networks have dominated the field of volumetric image segmentation due to their flexibility, optimized modular design, and success in all medical image modalities \citep{azad2022medical}, including convolution-based \citep{hesamian2019deep} and Transformer-based networks \citep{shamshad2023transformers}. Benefiting from the strong representation and generalization ability of deep models, U-shape networks \citep{ronneberger2015u} can easily handle the segmentation task of various datasets, especially for anatomies with clear boundaries. However, for some challenging datasets in which there exist blurred boundaries between different anatomical structures, UNet-based networks will suffer from poor predictions with inter-class boundary confusion due to the lack of boundary shape constraints \citep{wang2022boundary}. Datasets with different characteristics may all result in confused boundary segmentation. Here in our work, according to the intrinsic properties of datasets, we group these datasets into three categories:

\begin{itemize}

\item \textbf{Unclear boundaries or noise near boundaries}: Some datasets present an indistinct boundary contrast between different classes of objects, which poses a significant challenge to human vision perception of locating edges. In the meantime, deep neural networks encounter the same trouble in the segmentation task. For instance, boundaries between the lobes of the right lung are blurred \citep{park2020fully}, which are even invisible in some computed tomography (CT) scans. Furthermore, some patients infected with COVID-19 exhibit patchy shadows on lung lobes in CT scans \citep{xie2020relational}. The artifacts and random noise beside boundaries will bring uncertainty to deep models, which substantially increases the difficulty of precise boundary segmentation. 

\begin{figure}[!t]
\centerline{\includegraphics[width=0.8\linewidth]{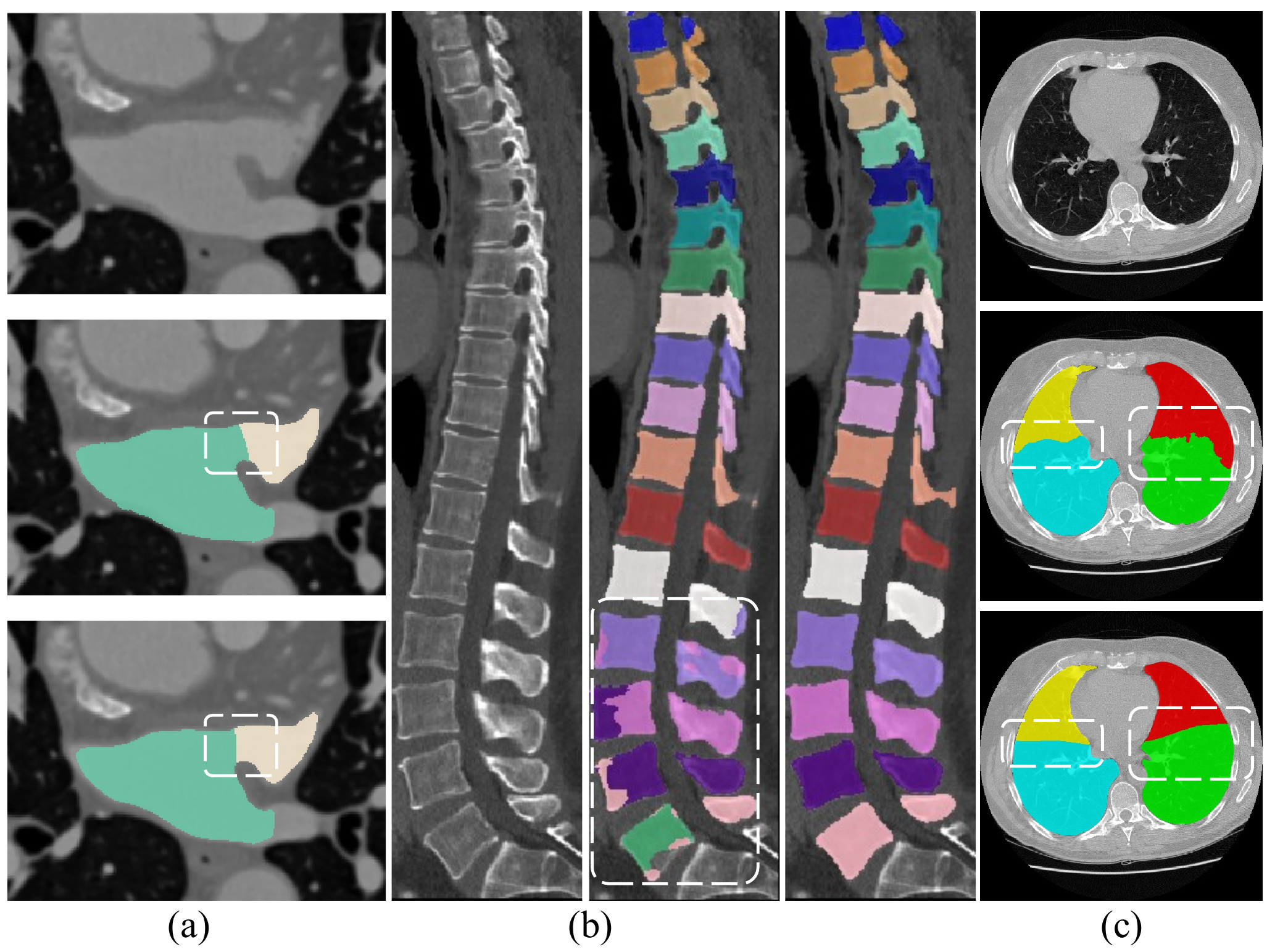}}
\caption{Three types of boundary confusion. (a) The obstacle in obtaining the unified standard for annotations of inter-class boundaries (Upper and lower annotations from junior and senior clinicians respectively). (b) Adjacent boundaries between anatomies with similar appearances but different classes. (From left to right: image/ prediction by vanilla 3D UNet/ground truth). (c) Unclear boundaries or noise near boundaries. (From top to bottom: image/ prediction by vanilla 3D UNet/ground truth).}
\label{boundary confusion}
\end{figure}

\item \textbf{The obstacle in acquiring a uniform standard for annotations of inter-class boundaries due to skill variations}: For some anatomical structures closely connected to each other, it is very significant but challenging to have an accurate segmentation mask of intricate interfaces between different anatomies. However, for structures like the aortic lumen and wall \citep{wang2017segmentation}, the left atrial appendage (LAA) and left atrium (LA) \citep{zheng2013multi} and so forth, there is a dilemma in achieving an identical standard for annotating boundaries between two anatomies. As illustrated in Figure \ref{boundary confusion}, upper and lower annotations are carried out by junior and senior clinicians. Here a lack of uniform standards for boundary ground truths (GTs) will also introduce 
 uncertainty to the output of neural networks \citep{song2022learning}, which finally results in the phenomenon of boundary confusion.


\item \textbf{Adjacent boundaries between anatomies with similar appearances but different classes}: if anatomies show homogeneous structures, there is likely to be segmentation inconsistency inside each anatomical structure \citep{you2023verteformer}. Concretely, each anatomical structure should be labeled as a unique class. However, the same label will occur inside other anatomies with a homogeneous appearance, causing inaccurate boundary segmentation for anatomical structures. This phenomenon also belongs to the scope of boundary confusion. As shown in Figure \ref{boundary confusion}, typical examples include vertebral CT images \citep{sekuboyina2021verse}. Besides, When objects are close to each other but not intersecting, models may falsely fill in the blank space between objects if no additional constraints are added to train the segmentation networks \citep{gupta2022learning}. 

\end{itemize}

Existing methods on refining the boundary segmentation and addressing the boundary confusion can be grouped into three categories. \ding{182} The first strategy attempts to exert a strong loss constraint on boundaries via the multi-task learning paradigm \citep{wang2022boundary, hatamizadeh2019end}. Furthermore, some loss functions \citep{kervadec2021boundary, karimi2019reducing, ma2021loss} are specifically devised for enhancing models' perception abilities for the boundary region. However, the loss regularization on boundaries is expert in segmenting objects with relatively fixed shapes. For anatomical structures with irregular shapes, a hard shape constraint will be detrimental to models' generalization abilities \citep{akbari2021does, zhang2021understanding}, thus degrading the segmentation performance. Moreover, for datasets without a uniform golden standard for boundaries, uncertain supervision information will also affect models' representation abilities for uncertain boundaries. \ding{183} Then some researchers propose complex post-processing procedures to refine coarse boundary predictions, such as \citep{yuan2020segfix, kirillov2020pointrend, zhou2020deepstrip, tang2021look, zhu2023adaptive} etc. Nonetheless, all the methods above mainly emphasize perfecting predicted masks for high-resolution and high-quality images with distinct boundary contrast, and are not applicable to ambiguous or unclear boundaries in medical domains \citep{xie2022uncertainty}. Besides, most existing studies propose an additional framework to
refine coarse predictions to fine predictions, which is not an end-to-end pipeline and will incur larger computational costs. Thus, this type of strategy also cannot effectively and efficiently address the aforementioned inter-class boundary confusion. \ding{184} Instead, the third strategy could work to some degree, with the mechanism of enhancing features representing uncertain boundaries via introducing prior information. However, these attention-based methods are mainly devised for specific anatomies with only one type of boundary confusion and deficient of generalization properties for various medical datasets.

In fact, existing methods on strengthening boundary features are still not skilled in dealing with boundary confusion. Here we give a theoretical explanation. Inter-class boundaries exist as individual objects. Previous methods essentially enhance the boundary itself, which is a slender structure with a width of one voxel. However, neural networks are indeed not skilled at localizing and segmenting tiny objects \citep{hesamian2019deep}. Thus, three types of methods mentioned above cannot finely address the precise segmentation of uncertain boundaries on the ground that they overemphasize this small structure. And the boundary tends to be a continuous interface. A slight turbulence will result in continuously wrong predictions as we can see from the structure of lung lobes and LA/LAA in Figure \ref{boundary confusion}. Besides, for multiple vertebrae with continuous labels, an inaccurate boundary prediction for a singular vertebra will result in incorrect predictions for a sequence of vertebrae, which should be strictly avoided in the process of image-guided analysis and diagnosis.

\begin{figure}[!t]
\centerline{\includegraphics[width=0.85\linewidth]{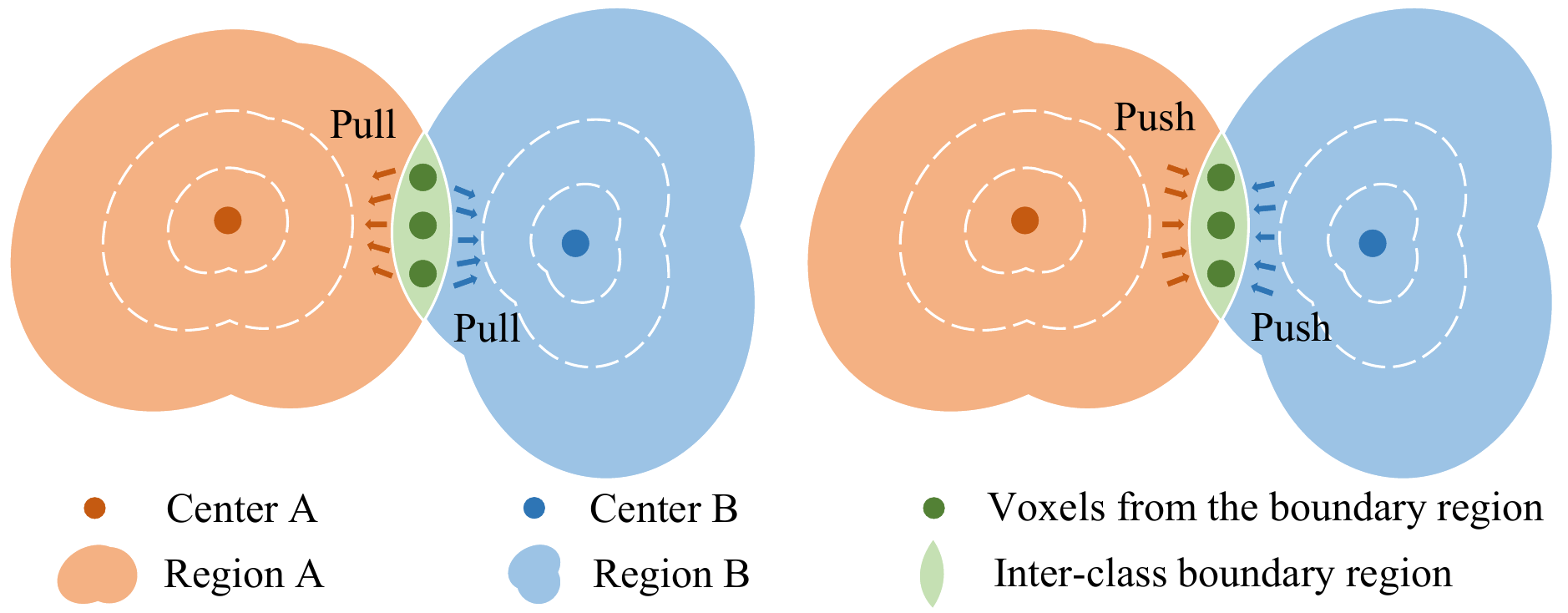}}
\caption{Two adversarial dynamics acting on boundaries. The boundary region is stretched by the pulling force, which is produced by clustered class centers. The boundary region is squeezed by the pushing force, which derives from the diffusion effect. The realistic boundary exists inside the intersected region, which implies boundary uncertainty. Pulling and pushing dynamics will enlarge and reduce that uncertainty.}
\label{boundary pull-push}
\end{figure}

The logic of previous methods indicates that the boundary itself is a natural byproduct, independent of the foregrounds. Different from that, we give a new perspective on the mechanism of boundary generation. Concretely, boundaries are generated with the dynamic interactions with two adjacent regions, including the pulling force from boundaries to drive these two regions close to each other and the pushing force to set them apart. According to Newton's Third Law, boundaries are squeezed and stretched under the effects of reactionary pulling and pushing forces as depicted in Figure \ref{boundary pull-push}. Here the inter-class boundary region implies boundary uncertainty, which can be amplified and reduced under the influence of pulling and pushing dynamics from adjacent regions. After a dynamic equilibrium, the intersected region will form the final interface between two adjacent anatomies.

Based on this theory, we propose an adversarial pull-push mechanism for medical image segmentation with inter-class boundary confusion. The whole network termed PnPNet, is mainly composed of two parts, the Pushing and Pulling branches in respective, which provide pushing and pulling forces for the uncertain boundary region. And these two forces will reach a state of dynamic equilibrium during training for a more precise boundary prediction. For the pushing branch, we follow the structural design of our previous conference work \citep{you2023semantic}. Motivated by the diffusion theory, we propose a semantic guidance module based on differential operators to refine features from the ambiguous boundary region, which is called the semantic difference module (SDM). Here we introduce semantic information from deeper layers to guide the diffusion process. However, the previously proposed SDM lacks the constraint for the difference kernel, in which all values might be negative and the difference information no longer exists. Thus, we improve SDM by introducing Explicit and Implicit Differential relations into the kernel, called EID kernel. Specifically, explicit differential relations are built up by setting vertexes of the $3 \times 3 \times 3$ cube as fixed positive and negative values, and implicit relations are produced by means of the remaining 19 values in the kernel. The improved SDM provides the pushing force for intersected boundary regions under diffusion guidance, which attempts to shrink the boundary uncertainty. Nevertheless, networks with only the single-directional pushing force cannot exquisitely model the uncertainty of boundaries, which is not feasible for cases plagued by boundary
confusion. As a result, uncertain boundaries cannot be precisely located, and that is why we aim to introduce the pulling branch in our work.

We attempt to stretch the inter-class boundary region via the pulling branch. Motivated by the structure of Maskformer and Mask2former \citep{cheng2021per, cheng2022masked}, we introduce a sequence of learnable object queries, which can be regarded as clustered centers of objects with different classes. Here clustered segmentation masks are attained by calculating the similarity between clustered centers and voxel-wise embeddings from skipped features \citep{ronneberger2015u}. And clustered class centers are iteratively updated via the interaction between skipped features and clustered segmentation masks. In the iterative optimization process, class centers move along the gradient direction of iso-surfaces in Figure \ref{boundary pull-push}, till the center ground truth. In the meantime, both sides of the boundary region are deformably spread toward different class centers via the pulling force. Here the core component of the pulling branch is referred to as the class clustering module (CCM), which is complementary to SDM. Furthermore, to accomplish a better convergence of learnable class centers, the center atlas and semantic prior information are introduced to generate high-quality initial class centers. The proposed SDM and CCM bear the plug-and-play characteristic and can be plugged into different neural networks to enhance the segmentation performance for cases with confused boundary predictions. In conclusion, our work contains the following contributions:

1) We summarize three types of dataset characteristics leading to inter-class boundary confusion in medical image segmentation and propose PnPNet to address this problem by modeling the dynamics of the interaction between the intersected boundary region and its adjacent regions.

2) We introduce the pushing branch as a pushing force to squeeze the boundary region. In this branch, the semantic difference module (SDM) from our previous work is improved by introducing explicit and implicit differential relations, and it bears stronger representation abilities for confused boundaries.

3) We propose the pulling branch as a pulling force to stretch the uncertain boundary region. In this branch, the class clustering module (CCM) is utilized for achieving clustered segmentation masks and iterative updates for class centers. Besides, segmentation performance highly depends on initial class centers. Thus, we facilitate models' learning by generating a high-quality center atlas and exerting semantic prior information into centers.

4) Our proposed model outperforms other convolutional neural networks (CNNs) and Transformer-based models on four challenging datasets. Furthermore, extensive experimental results demonstrate that SDM and CCM can boost models' segmentation performance as plug-and-play modules.

\section{Related Works}

\subsection{Methods on Refining Boundary Segmentation}
Boosting the precise localization of boundaries has been one of the most studied scenarios in the field of medical image segmentation. We categorized previous works on this topic into three types, including specifically designed loss functions introduced via the multi-task learning paradigm, post-processing strategies, and attention-based enhancement modules.

\textbf{Multi-task learning paradigm joined with specific loss functions}: Previous studies \citep{hatamizadeh2019end, zhang2019net, chen2020supervised, wang2022eanet, wang2022boundary} tend to enhance edge feature representations via the introduction of boundary regularization besides the segmentation loss. Zhang et al. \citep{zhang2019net} proposed an edge guidance module to learn edge attention representations in the early encoding layers, which are then transferred to the multi-scale decoding layers. Chen et al. \citep{chen2020supervised} addressed the instance segmentation task by devising the edge attention module to highlight objects and suppress background noise. EANet \citep{wang2022eanet} provides an iterative edge attention network to generate more accurate saliency maps. Besides, some boundary-aware loss terms \citep{kervadec2021boundary, karimi2019reducing, ma2021loss, wang2022active, borse2021inverseform} are present to enhance models' perception abilities for boundaries. Borse et al. \citep{borse2021inverseform} proposed a boundary distance-based measure, InverseForm, which is more capable of capturing spatial boundary
transforms than cross-entropy-based measures, thus resulting in more accurate segmentation results. Hausdorff Distance loss \citep{karimi2019reducing} is devised to reduce the Hausdorff Distance between boundaries of predictions and ground truth masks. Kervadec et al. \citep{kervadec2021boundary} designed a boundary loss, which takes the form of a distance metric on the space of contours. This loss mitigates the challenge of segmenting highly unbalanced anatomies since it uses integrals over the interface between regions. The active boundary loss \citep{wang2022active} is proposed by formulating the boundary alignment problem as the differentiable direction vector prediction, to guide the movement of predicted boundaries. However, these boundary regularizations mentioned above cannot effectively address the segmentation for anatomies with irregular shapes.

\textbf{Post-processing procedures}: Post-processing strategies are usually utilized to refine the shape of coarse boundary predictions from the first-stage network. PointRend \citep{kirillov2020pointrend} introduced a module that performs point-based segmentation predictions at adaptively selected locations based on an iterative subdivision algorithm. SegFix \citep{yuan2020segfix} proposed to replace originally unreliable predictions of boundary pixels with predictions of interior pixels. And this approach builds the correspondence by learning a direction away from the boundary pixel to an interior pixel. Tang et al. \citep{tang2021look} extracted and refined a series of small boundary patches along the predicted instance boundaries. Huynh et al. \citep{huynh2021progressive} devised a multi-scale framework that has multiple processing stages, where each stage corresponds to a magnification level. SharpContour \citep{zhu2022sharpcontour} designed a novel contour evolution process together with an instance-aware point classifier. This method deforms the contour iteratively by updating offsets in a discrete manner. APPNet \citep{zhu2023adaptive} developed the global-local aggregation module to model the context between global and local predictions. It also introduced an adaptive point replacement module to compensate for the lack of fine detail in global prediction and the overconfidence in local predictions. Nevertheless, the methods proposed above require high-quality images with discriminative boundaries, and are not effective in tackling uncertain boundaries in medical image scenarios. Additionally, these methods are specifically designed to refine the boundary prediction of individual objects, not applicable to inter-class boundaries.

\textbf{Implicit enhancement modules}: These methods \citep{fan2020pranet, lee2020structure, cao2022novel, xie2022uncertainty, gupta2022learning, lin2023rethinking, wang2023xbound, yang2023ept} function well for segmentation tasks on specific datasets by exerting shape priors of boundaries into neural networks. Lee et al. \citep{lee2020structure} proposed a novel boundary-preserving block (BPB) with the ground-truth structure information indicated by experts. Xie et al. \citep{xie2022uncertainty} used the confidence map to evaluate the uncertainty of each pixel to enhance the segmentation of ambiguous boundaries. Besides, a boundary attention module (BAM) \citep{cao2022novel} is introduced to excavate abundant boundary features. BAM is hierarchically introduced into UNets to highlight outline information of foregrounds. PraNet \citep{fan2020pranet} devised the reverse attention (RA) module to mine the boundary cues, and this module is able to establish the relationship between areas and boundary cues. Saumya Gupta et al. \citep{gupta2022learning} introduced a novel topological interaction module to encode the topological interactions into a deep neural network, especially for boundary voxels. Yi Lin et al. \citep{lin2023rethinking} proposed a dedicated boundary detection operator to enhance the learning capacity on the boundary region. XBound-Former \citep{wang2023xbound} proposed the implicit, explicit and cross-scale boundary learner to catch boundary knowledge. In this work, we propose an adversarial pull-push mechanism by modeling the interaction between the boundary and its adjacent regions, instead of regarding the boundary as an individual and slender structure as in previous studies.

\subsection{Preliminaries on diffusion theory}
Diffusion is a physical phenomenon, in which molecules spread from regions with higher concentrations toward regions with lower concentrations \citep{sapiro2006geometric, tan2022semantic}. Then the whole system tends to be balanced. For a feature vector $F$ to be smoothed, the diffusion process can be modeled as the following partial differential equation:
\begin{eqnarray}
  & \frac{\partial F}{\partial t} = D \cdot \nabla^{2} F
  \label{diffusion}
\end{eqnarray}

where $D$ is the diffusivity function determining the diffusion speed along each direction, $\nabla$ is the gradient operator. In our application, the stable state of $F(t)$ will present a more accurate localization for inter-class boundaries.

Linear isotropic diffusion ($D$ is equal to a constant) cannot be applied to complex scenes because the diffusion velocity is the same in all directions. For a spatial-dependent function $D = D(x, y, z)$, the process is linear anisotropic. However, if we aim to extract refined boundary features, adopting linear diffusion processes will smooth both backgrounds and the edges. A more feasible solution is to devise complex diffusion functions $D = D(F)$ with nonlinear characteristics \citep{weickert1999coherence, weickert1998efficient}. As a result, the diffusion process exerts more smoothing to regions parallel to boundaries compared to regions vertical to these edges.

Detailedly, given a feature $F$ where regions of uncertain boundaries are not highlighted, it is updated by the diffusion process in infinite time. The diffusion adjacent to confused boundaries should be restrained, while the diffusion far away from boundaries is promoted. And the final state of the diffused feature will accurately localize boundaries between different anatomies.

\section{Methodology}
\subsection{Overview}
As illustrated by Figure \ref{The whole structure}, PnPNet consists of the baseline encoder-decoder structure, the semantic difference module (SDM), and the class clustering module (CCM), in which SDM and CCM can be plugged into different CNNs, Transformer-based, and MLP-based models, including 3D UNet \citep{cciccek20163d}, nnUNet \citep{isensee2021nnu}, MedNeXt \citep{roy2023mednext}, TransUNet \citep{chen2021transunet}, Swin UNETR \citep{tang2022self}, and UNeXt \citep{valanarasu2022unext} etc. Detailedly, SDM will generate a pushing dynamic to compress the inter-class boundary region, while CCM can provide a pulling dynamic to expand that region. During training, these two adversarial forces generate mutual constraints and reach a state of dynamic equilibrium after convergence. Thus, an elaborate design by combining SDM and CCM will facilitate to achieve finer boundary predictions. In the following subsections, we will detailedly discuss these two modules and the interactive style between pushing and pulling tokens.

\subsection{Pushing Branch}
\begin{figure*}[!t]
\centerline{\includegraphics[width=0.95\linewidth]{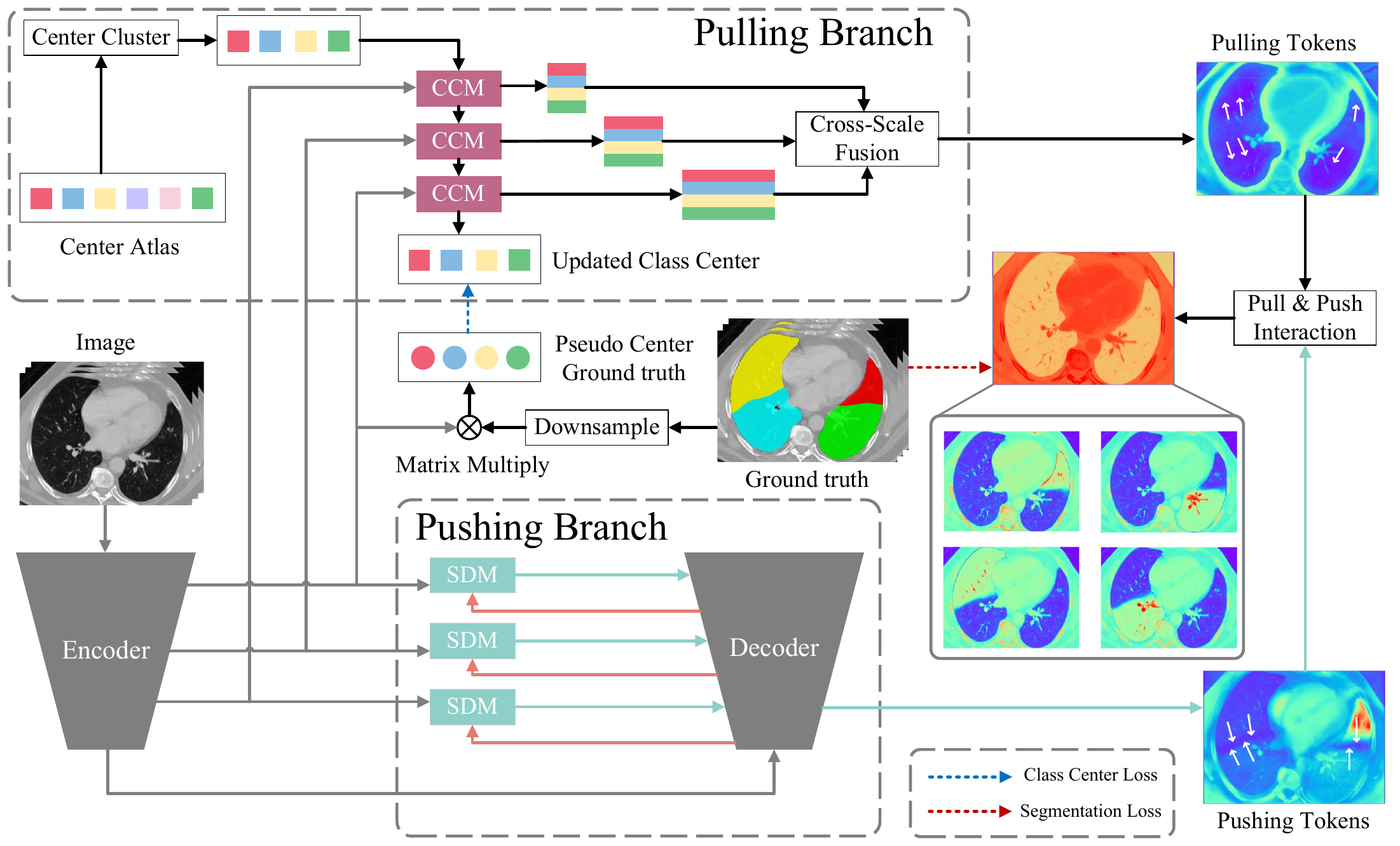}}
\caption{The whole network architecture is based on the pull-push mechanism. (a) For the pushing branch, the semantic difference module (SDM) and convolutional decoder are employed to squeeze the boundary region. (b) For the pulling branch, we respectively introduce N class centers as the average representation metric for each class. And the class clustering module (CCM) is devised to stretch the boundary region. Multi-scale mask embeddings are fused into pushing tokens. Besides, the center atlas and pseudo-center GT are employed to relieve the convergence challenge of class centers. (c) Pushing and pulling mask tokens are interacted into decoded class tokens.}
\label{The whole structure}
\end{figure*}

To compress the boundary region between anatomies with different classes, we propose the semantic difference module (SDM) as a pulling force to reduce the boundary uncertainty. Motivated by the diffusion process, we formulate the process of enhancing boundary feature representations by solving a second-order partial differential equation. Boundary features are highlighted under the diffusion guidance, in which the diffusion process close to the inter-class boundary region is restrained and far from those is facilitated. Thus, the diffusion effect can be considered as a type of pushing force generated from two adjacent anatomies, which can squeeze the inter-class boundary region to some extent as revealed in Figure \ref{The whole structure}. In that case, raw skipped features are refined as boundary-compressed features. Finally, we utilize the convolutional decoder to fuse these boundary-enhanced skipped features in the Upsample-Concatenate-Convolution way. For more details about the specific decoder structure, please refer to our released source code \url{https://github.com/AlexYouXin/PnPNet}.

\subsubsection{Semantic Difference Module}
\begin{figure*}[!t]
\centerline{\includegraphics[width=0.95\linewidth]{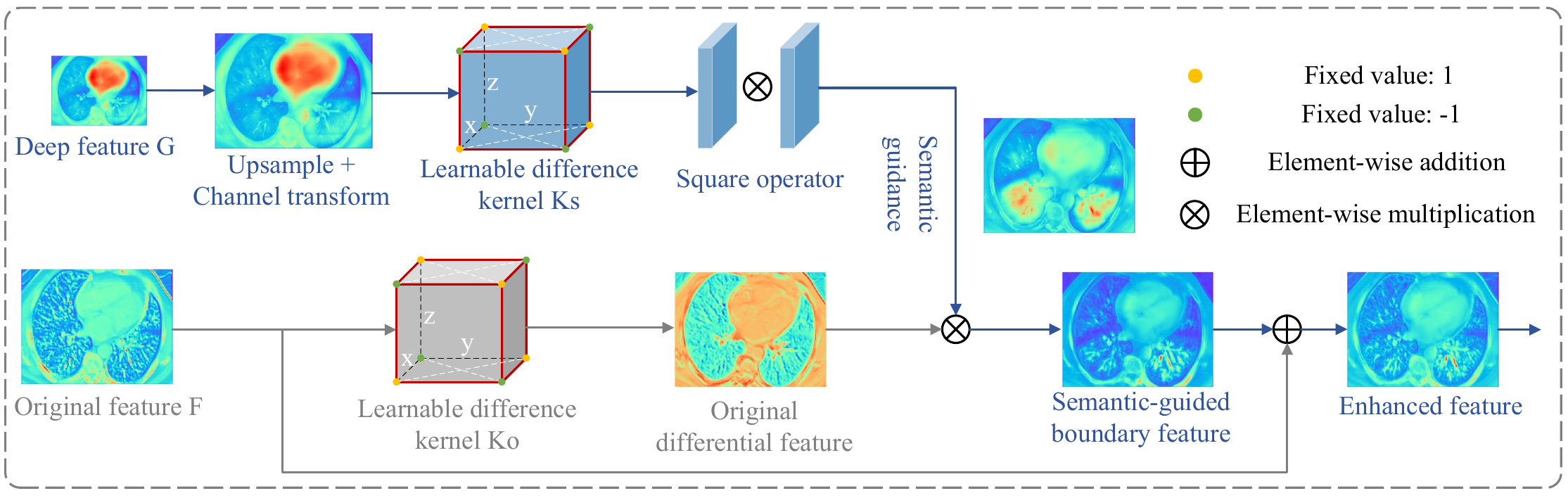}}
\caption{The detailed structure of SDM. The original differential feature is generated by the differential operation on the raw skipped feature F. Then semantic guidance from the deep feature G is introduced to refine the original differential feature as the enhanced feature.}
\label{SDM structure}
\end{figure*}

Since semantic information is required to guide the localization of uncertain boundaries, we introduce the deep feature $G$ from the precedent decoder layer into the diffusivity function. Here we adopt the differential map of deep features $\nabla \textit{G}$ as the semantic guidance map. And the square term $h(|\nabla \textit{G}|^2)$ is deployed as function $D$ to model nonlinear characteristics of the diffusion process, where $h$ is a convolutional projection function. In terms of \citep{sapiro2006geometric}, Eq. \ref{diffusion} can be approximately solved via iterative updates as depicted by the following equations:
\begin{eqnarray}
  & \hat{F}^{t+1}_{\boldsymbol p} = \displaystyle \sum_{\widetilde{\boldsymbol p} \in \delta_{\boldsymbol p}}{h(|G_{\widetilde{\boldsymbol p}} - G_{\boldsymbol p}|^2) \cdot (F^{t}_{\widetilde{\boldsymbol p}} - F^{t}_{\boldsymbol p})} \\
  & F^{t+1}_{\boldsymbol p} = \lambda \cdot F^{t}_{\boldsymbol p} + \nu \cdot \hat{F}^{t+1}_{\boldsymbol p}
  \label{iterative solution}
\end{eqnarray}

\noindent where $\boldsymbol p$ is the index of feature maps, $\delta_{\boldsymbol p}$ is the $3 \times 3 \times 3$ local neighborhood centered at $\boldsymbol p$, $\lambda$ and $\nu$ are weighting coefficients. Indeed, $F^{t}_{\widetilde{\boldsymbol p}} - F^{t}_{\boldsymbol p}$ is the differential information of the original feature $F^{t}$ at point $\boldsymbol p$, representing abundant boundary information, which contains complicated boundary features of anatomies as shown in Figure \ref{SDM structure}. However, predicted inter-class boundaries are not accurate enough only with the diffusion process. Thus, the semantic difference guidance $|G_{\widetilde{\boldsymbol p}} - G_{\boldsymbol p}|^2$ is introduced to generate refined boundary feature $\hat{F}^{t+1}$. $\hat{F}^{t+1}$ will diffuse into the stable state as $t$ increases, which can highlight boundaries between different classes, and suppress the activation on other irrelevant boundary regions. And refined feature $F^{t+1}$ is attained by fusing the original feature $F^{t}$ with enhanced boundary feature $\hat{F}^{t+1}$.

We design the semantic difference module based on Eq. \ref{iterative solution} as illustrated by Figure \ref{SDM structure}. Motivated by the fact that there exists an anisotropic distribution for various medical datasets in $x$, $y$ and $z$ dimensions, traditional edge operators cannot finely extract the differential map of feature $F$. Therefore, we propose a learnable boundary operator, which bears different values in each position of the kernel. In our previous work, we fixed the center value as $-1$ to maintain the difference attribute of the edge kernel. However, there is a lack of a constraint on other non-centering values. If the values in kernels are all negative, then the filter will be a low-pass blurring filter producing a negative output.

\subsubsection{Explicit-Implicit Difference Kernel}
To address this limitation, we make a further improvement by introducing \textbf{E}xplicit and \textbf{I}mplicit \textbf{D}ifferential information into the learnable \textbf{Kernel}, termed EID kernel. Specifically, 8 vertexes in the $3 \times 3 \times 3$ cube are set as fixed values. According to Figure \ref{SDM structure}, yellow and green vertexes are set as $1$ and $-1$ respectively, and adjacent vertexes must be a pair of $(-1, 1)$. Consequently, the differential filter bears 12 edge-oriented differential information, which is explicit. Besides, the remaining 19 values in the filter are set as learnable parameters, which can produce implicit differential relations inside the kernel. The revised description of the enhanced boundary feature is calculated by the following formulas.
\begin{eqnarray}
  & \hat{F}^{t+1}_{\boldsymbol p} = \displaystyle \sum_{\widetilde{\boldsymbol p} \in \delta_{\boldsymbol p}}{\omega_{\widetilde{\boldsymbol p}} \cdot |\alpha_{\widetilde{\boldsymbol p}} G_{\widetilde{\boldsymbol p}} - G_{\boldsymbol p}|^2 \cdot (\beta_{\widetilde{\boldsymbol p}} F^{t}_{\widetilde{\boldsymbol p}} - F^{t}_{\boldsymbol p})}
  \label{revised iterative solution}
\end{eqnarray}
\begin{equation}
    \alpha_{\widetilde{\boldsymbol p}}, \beta_{\widetilde{\boldsymbol p}} =
    \begin{cases}
        -1, & \text{if} \quad \widetilde{\boldsymbol p} \in  \mathcal{S}_{g} \\
        1, & \text{else if} \quad \widetilde{\boldsymbol p} \in  \mathcal{S}_{y} \\
        x, & \text{else}
    \end{cases}
      \label{revised kernel}
\end{equation}

Where $\alpha$ and $\beta$ refer to different learnable edge operators for feature $F$ and semantic feature $G$ respectively, $\mathcal{S}_{g}$ and $\mathcal{S}_{y}$ represent the set of green and yellow vertexes in the cube. And $\omega_{\widetilde{\boldsymbol p}}$ means a vanilla $3 \times 3 \times 3$ convolution kernel.

In conclusion, the semantic difference module can better localize boundaries between two anatomies under diffusion guidance, in which the EID kernel bears explicit and implicit differential information. And SDM serves as a pushing force to squeeze the boundary region, then shrinks the boundary uncertainty. The decoder structure in the pushing branch is deployed to fuse these enhanced skipped features with the boundary region compressed. However, only the pushing branch will give a certain prediction for boundaries, which is not suitable for cases suffering from boundary confusion. And that is why the pulling branch is required to introduce the uncertainty into the inter-class boundary region.

\subsection{Pulling Branch}
To generate an adversarial force that stretches the inter-class boundary region, we propose the class clustering module (CCM), serving as a pulling force. In the structure of Maskformer \citep{cheng2021per} and Mask2former \citep{cheng2022masked}, learnable object queries are employed to model semantic characteristics of each instance object. Here we reformulate object queries as class centers. Motivated by K-means clustering \citep{hartigan1979algorithm}, we iteratively update class centers and mask embeddings for objects with different semantics in the pulling branch. Here mask embeddings can be deployed to calculate clustered segmentation masks, thus driving the feature space of each semantic class tighter. Moreover, class centers correspond to each semantic class in the semantic segmentation task. In the iterative optimization process, class centers move towards the center ground truth. In that way, both sides of the boundary region
are expanded toward different class centers under the guidance of the pulling force from class centers. Thus, CCM can be practical to stretch the inter-class boundary region.

\begin{figure}[!t]
\centerline{\includegraphics[width=0.95\linewidth]{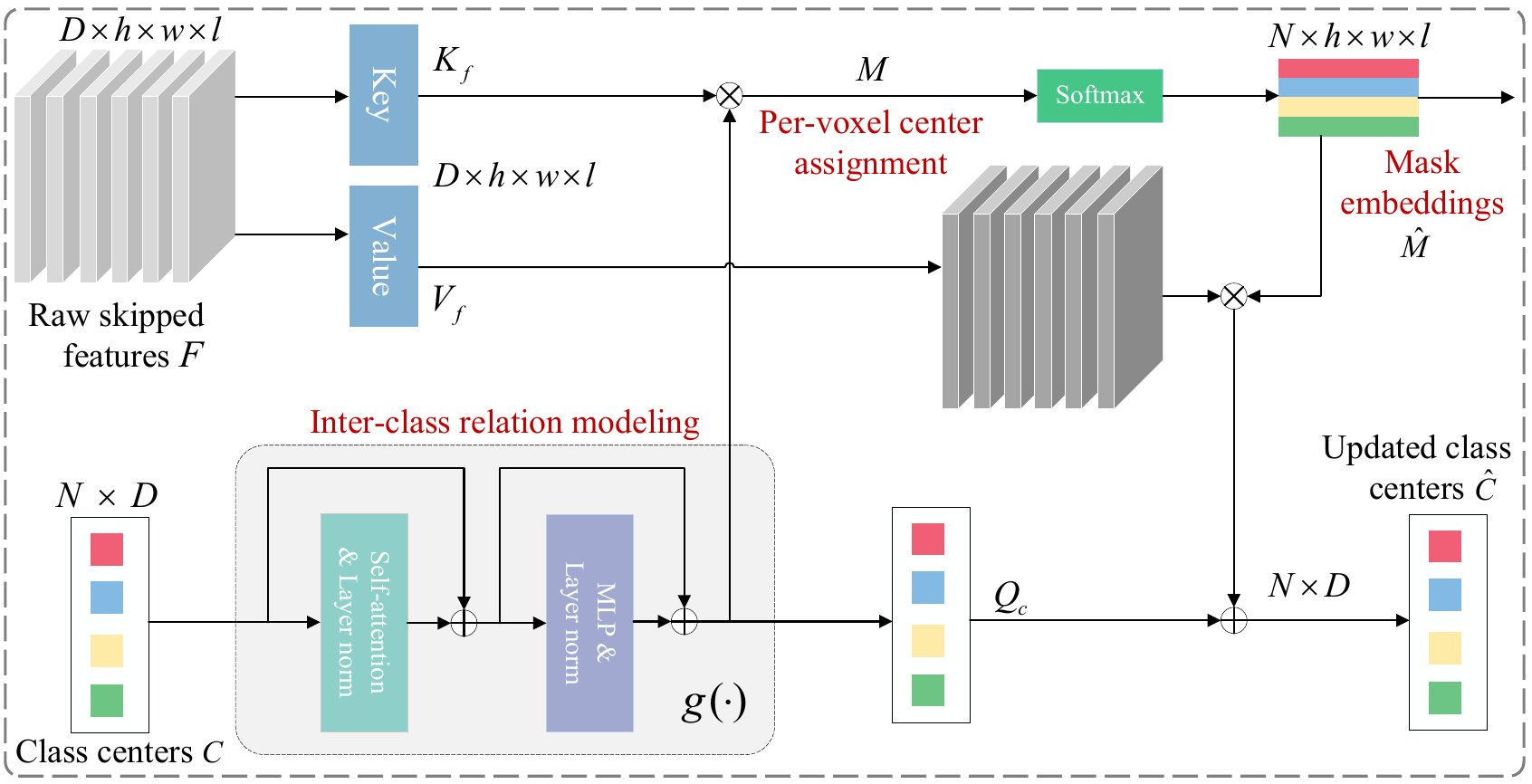}}
\caption{The detailed structure of CCM. Mask embeddings are achieved by calculating the voxel-wise similarity with each class center. After the Softmax operation, mask embeddings are normalized across different channels. Then exerting the voxel-wise class information on the transformed skipped feature, we can attain average class embeddings, which are employed to update class centers.}
\label{CCM structure}
\end{figure}

Specifically, the mechanism of CCM consists of two aspects. One is the generation process of mask embeddings $\hat{M}$. As depicted by Figure \ref{CCM structure}, mask embeddings $M$ are achieved by calculating the voxel-wise similarity between raw skipped feature embeddings $F$ and class centers $C$. Here centers exist as N D-dimension vectors ($D$ is set as 192, $N$ refers to the number of segmentation classes). And the dimension of F is set as $D \times h \times w \times l$, in which ($h$, $w$, $l$) accounts for the ratio k of the input size ($H$, $W$, $L$), with k equal to ($\tfrac{1}{2}$, $\tfrac{1}{4}$, $\tfrac{1}{8}$). By carrying out the Softmax operation, we normalize mask embeddings $M$ across different channels as $\hat{M}$. Finally, the channel index of the maximal per-voxel activation value corresponds to the center assignment result. The whole calculation can be modeled as the following equation:
\begin{eqnarray}
  & \hat{M} = \mathop{Softmax} \limits_{n: 1 \Rightarrow N} (M) \\
  & M(n, x, y, z) = \displaystyle \sum_{i=1}^{D} (Q_{c}(n, i) \cdot K_{f}(i, x, y, z)) \\
  & Q_{c} = g(C)
  \label{mask embeddings}
\end{eqnarray}

$n$ means the channel index of mask embeddings $M$, ($x$, $y$, $z$) is the voxel index, $K_{f}$ represents the key vector after a convolutional transform on $F$, and $g(\cdot)$ refers to combinational operations of self-attention, layer normalization (LN), residual connection, and Multi-Layer Perceptron (MLP) \citep{dosovitskiy2020image}. Here $g(\cdot)$ is aimed at modeling the relations between different class centers by means of the self-attention mechanism. Due to the fact that the number of patch tokens is $N$, this operation only requires $\mathcal{O}(N^{2}D)$ computational complexity, which is a constant value.


The other aspect is the updating process of class centers. Following the algorithm of K-means, we can attain average class embeddings by exerting the voxel-wise class information on the transformed value vector $V_{f}$. And the average representations serve as the incremental to update class centers. $\hat{C}$ refers to updated class centers. Thus, the whole process can be summarized as follows:
\begin{eqnarray}
  & \hat{C} = Q_{c} + \displaystyle \sum_{x=1}^{h} \sum_{y=1}^{w} \sum_{z=1}^{l} \hat{M}(:, x, y, z) \cdot V_{f}^T(x, y, z, :)
  \label{center update}
\end{eqnarray}

Owing to the fact that the self-attention mechanism bears the low-pass property \citep{park2021vision}, CCM is expert in aggregating information of local regions with the specific class, then performing clustering effects. However, similar to DETR \citep{carion2020end}, the structure of maskformer and its relevant architectures \citep{yu2022cmt, cheng2022masked} face the same convergence challenge, which is well-trained class centers. Indeed, a faster and better convergence highly depends on a set of high-quality class centers \citep{meng2021conditional, liu2021dab}. Here class centers will interact with skipped features to generate final segmentation masks, and are closely related to objects' semantic information. Therefore, it is essential to facilitate models' learning by generating high-quality initial class centers and exerting semantic prior information into centers.

\subsubsection{Center Atlas}
Firstly, we make an improvement to the CCM structure from the perspective of generating high-quality class centers. For the semantic segmentation task, class centers bear the same number $N$ as the segmentation classes. And each center reflects the characteristics of a specific class. However, these centers cannot cover the property of the whole dataset for the sake of various data distributions, including the diversity of shape and texture. To some extent, fitting the characteristics of datasets with less learnable centers will further enlarge the convergence difficulty.

Thus, we introduce the center atlas, which contains $\hat{N}$ referenced centers for $N$ semantic classes ($\hat{N} > N$). In the atlas, a group of adjacent centers contribute to the specific class. These centers representing a certain class can better model data distributions of class-wise objects, which is significant for extracting semantic information from skipped features. Thus relational centers are clustered as an 
initial class center with linear projection layers and GELU activation function, then the center atlas is merged as $N$ initial class centers. Theoretically, clustered class centers can be more precisely measured as average metric representations of semantic classes. Therefore, the inter-class distance \citep{zhou2022rethinking} between different class centers can be maximized, and the intersected boundary region can be further extended.

Besides, according to the concept of the interactive segmentation \citep{lin2016scribblesup}, the center atlas is equivalent to n prompting points. As a foundation model, SAM \citep{kirillov2023segment} demonstrates that prompt-based techniques will have broader applications in natural scenes. Many recent researches on Medical SAM \citep{huang2023segment, mazurowski2023segment, cheng2023sam} further validate this perspective that point and bounding box prompts are beneficial to improve the segmentation performance. Here $M$ referenced centers in the atlas can be viewed as positive point embeddings, which represent the average metric for objects with different classes. In terms of the finding by \citep{huang2023segment, cheng2023sam}, more prompting points mean better model performance if the point number does not reach the saturation point. Thus, this is another strong proof of the effectiveness of the center atlas.

\subsubsection{Pseudo Labels for Class Centers}
To relieve the convergence difficulty of class centers, we attempt to introduce semantic priors into the pulling branch. Class centers implicitly characterize class-wise information, but we have no access to the ground truth. However, convolutional and Transformer-based features from the encoder have remarkable potentials to localize discriminative regions \citep{zhou2016learning, naseer2021intriguing}. That is to say, foreground areas will be highlighted in channel-wise features, although different channels tend to activate different objects. 
\begin{eqnarray}
  & \mathcal{C} = O \times F^{T} \\
  & \mathcal{L}_{cc} = \displaystyle \sum_{i=1}^{N} \sum_{j=1}^{D} ||\mathcal{C}_{ij} - \hat{C}_{ij}||^2
  \label{class center loss}
\end{eqnarray}
Thus, we can adopt the downsampled one-hot mask $O$ ($N \times \tfrac{H}{2} \times \tfrac{W}{2} \times \tfrac{L}{2}$) as weighted coefficients, to calculate the weighted sum with raw skipped features $F$ ($D \times \tfrac{H}{2} \times \tfrac{W}{2} \times \tfrac{L}{2}$) from the encoder as revealed in Figure \ref{The whole structure}. Here '$\times$' refers to the matrix multiplication. The final result $\mathcal{C}$ indicates class-wise feature embeddings ($N \times D$), which is indeed the average metric representation for different classes. $\mathcal{C}$ is utilized as the ground truth for updated class centers generated from the class clustering module close to the output layer. Here we choose the $L_{2}$ norm to calculate the class center loss $\mathcal{L}_{cc}$ as illustrated by Eq. \ref{class center loss}. By additionally introducing the regularized loss, we intend to inject semantic priors to guide the training for class centers.

\begin{table*}[!t]
  \begin{center}
  \caption{Comparison with other models on clean lung lobes. (LU, LL, RU, RM, RL: left upper, left lower, right upper, right middle, and right lower lobes, L: lobes of the left lung, R: lobes of the right lung, Mean: the average evaluation metric of all lobes. Bold numbers: the best, Underlined numbers: the second best.)}
  \label{tab clean}
  \resizebox{0.95\textwidth}{!}{
  \begin{tabular}{ccccccccccccc}
  \hline  
 
  \multirow{2}*{Method} & \multicolumn{3}{c}{LUL} & \multicolumn{3}{c}{LLL}  & \multicolumn{3}{c}{L} & \multicolumn{3}{c}{Mean} \\  
  \cmidrule(r){2-4}  \cmidrule(r){5-7}  \cmidrule(r){8-10}  \cmidrule(r){11-13}
  & Dice & HD95 & ASSD & Dice & HD95 & ASSD & Dice & HD95 & ASSD & Dice & HD95 & ASSD \\
  
  \hline  
  3D UNet \citep{cciccek20163d}  & 96.87 & 5.76 & 1.116
 & 96.65 & 6.90 & 1.124
 & 96.76 & 6.33 & 1.120
 & 93.80 & 8.83 & 1.860 \\
  VNet \citep{milletari2016v}  & 95.06 & 7.75 & 1.628
 & 94.93 & 7.70 & 1.417
 & 94.99 & 7.72 & 1.523
 & 93.68 & 8.56 & 1.851  \\
  ResUNet  \citep{diakogiannis2020resunet}   & 96.66 & 6.91 & 1.406
 & 96.67 & 5.85 & 1.023
 & 96.66 & 6.38 & 1.214
 & 94.11 & 7.45 & 1.674  \\
  TransUNet (3D)  \citep{chen2021transunet}  & 96.85 & 5.06 & 1.048
 & 96.76 & 5.04 & 0.950
 & 96.81 & 5.05 & 0.999
 & 94.35 & 7.10 & 1.490  \\
  TransBTS \citep{wang2021transbts}  & 89.69 & 19.49 & 4.446
 & 88.14 & 22.71 & 6.291
 & 88.91 & 21.10 & 5.369
 & 80.67 & 40.19 & 7.627  \\
  UNeXt  \citep{valanarasu2022unext}  & 96.15 & 6.02 & 0.978
 & 95.91 & 5.33 & 0.919
 & 96.03 & 5.67 & 0.949
 & 93.98 & 6.73 & 1.342  \\ 
  Swin UNETR \citep{tang2022self}  & 96.23 & 6.52 & 1.148
 & 96.20 & 6.30 & 1.030
 & 96.21 & 6.41 & 1.089
 & 94.21 & 7.29 & 1.436  \\
  3D UX-Net   \citep{lee20223d}  & 96.95 & 5.56 & \textbf{0.884}
 & 96.84 & 5.62 & 0.976
 & 96.90 & 5.59 & 0.930
 & 94.11 & 7.21 & 1.350  \\
  nnUNet \citep{isensee2021nnu} & 97.37 & 5.14 & 1.143
 & 97.34 & 4.37 & 0.784
 & 97.35 & 4.76 & 0.964
 & 94.72 & \underline{6.35} & 1.323  \\
  MedNeXt  \citep{roy2023mednext}  & \underline{97.41} & \underline{4.99} & 1.045
 & \underline{97.42} & \textbf{4.14} & \underline{0.769}
 & \underline{97.42} & \underline{4.57} & \underline{0.907}
 & \underline{94.94} & 6.46 & \underline{1.322}  \\
  Ours  & \textbf{97.55} & \textbf{4.87} & \underline{0.898}
 & \textbf{97.51} & \textbf{4.14} & \textbf{0.719}
 & \textbf{97.53} & \textbf{4.51} & \textbf{0.808}
 & \textbf{95.35} & \textbf{6.11} & \textbf{1.171}  \\
  \hline  
  \multirow{2}*{Method} & \multicolumn{3}{c}{RUL} & \multicolumn{3}{c}{RML}  & \multicolumn{3}{c}{RLL} & \multicolumn{3}{c}{R} \\  
  \cmidrule(r){2-4}  \cmidrule(r){5-7}  \cmidrule(r){8-10}  \cmidrule(r){11-13}
  & Dice & HD95 & ASSD & Dice & HD95 & ASSD & Dice & HD95 & ASSD & Dice & HD95 & ASSD \\
  
  \hline  
  3D UNet \citep{cciccek20163d}  & 93.24 & 12.51 & 2.266
 & 85.75 & 13.17 & 3.061
 & 96.51 & 5.82 & 1.731
 & 91.83 & 10.50 & 2.353 \\
  VNet \citep{milletari2016v}  & 94.39 & 6.51 & 1.683
 & 87.94 & 15.96 & 3.365
 & 96.09 & 4.87 & 1.161
 & 92.81 & 9.11 & 2.070 \\
  ResUNet  \citep{diakogiannis2020resunet}    & 93.90 & 7.62 & 1.598
 & 86.85 & 12.48 & 3.219
 & 96.46 & 4.41 & 1.124
 & 92.40 & 8.17 & 1.980 \\
  TransUNet (3D)  \citep{chen2021transunet}   & 94.22 & 6.55 & 1.469
 & 87.60 & 13.07 & 2.838
 & 96.33 & 5.77 & 1.148
 & 92.72 & 8.46 & 1.818 \\
  TransBTS \citep{wang2021transbts}   & 77.19 & 56.06 & 9.349
  & 62.41 & 21.33 & 6.421
 & 85.92 & 81.35 & 11.626
 & 75.17 & 52.92 & 9.132 \\
  UNeXt  \citep{valanarasu2022unext}   & 94.70 & 6.34 & 1.287
  & 86.60 & 12.42 & 2.730
 & 96.53 & \underline{3.57} & 0.793
 & 92.61 & 7.44 & 1.604 \\
  Swin UNETR \citep{tang2022self}  & 94.64 & 7.36 & 1.377
  & 87.54 & \underline{12.04} & 2.797
 & 96.46 & 4.24 & 0.830
 & 92.88 & 7.88 & 1.668 \\
  3D UX-Net   \citep{lee20223d}  & 94.59 & 6.68 & 1.306
 & 85.49 & 12.64 & \underline{2.564}
 & 96.67 & 5.53 & 1.021
 & 92.25 & 8.28 & 1.630 \\
  nnUNet \citep{isensee2021nnu}  & \underline{94.84} & \underline{6.11} & \underline{1.265}
 & 87.13 & 12.50 & 2.659
 & \underline{96.92} & 3.63 & \underline{0.762}
 & 92.96 & \underline{7.41} & \underline{1.562} \\
  MedNeXt  \citep{roy2023mednext}  & \underline{94.84} & 6.29 & 1.280
 & \underline{88.30} & 13.12 & 2.700
 & 96.73 & 3.77 & 0.819
 & \underline{93.29} & 7.73 & 1.600 \\
  Ours   & \textbf{95.80} & \textbf{6.09} & \textbf{1.007}
 & \textbf{88.99} & \textbf{11.96} & \textbf{2.519}
 & \textbf{96.93} & \textbf{3.49} & \textbf{0.714}
 & \textbf{93.90} & \textbf{7.18} & \textbf{1.413} \\

  \hline 
  \end{tabular}}
  \end{center}
\end{table*}

\subsection{Interaction between pushing and pulling branches}
These two branches generate pushing and pulling mask tokens in respective. For pushing tokens $\mathcal{T}_{push}$, the boundary region is squeezed into a narrow space. Besides, it is illustrated by Figure \ref{The whole structure} that pull tokens $\mathcal{T}_{pull}$ will expand the boundary area, which contains the interface of inter-class boundary GT. For the fusion way of tokens, since the convolutional decoder from the pushing branch shows more powerful capacities for shape representations \citep{isensee2021nnu, tang2022self}, an appropriate balance between two forces is significant to the training process Here we choose the pulling tokens as the main component to enhance the representation learning of the pulling branch, thus enlarge the boundary uncertainty. In that way, pushing masks serve as the auxiliary attention map for enhancing boundary features by means of the sigmoid function $\sigma$. Here we devise a simple fusion module as follows:
\begin{eqnarray}
  & \mathcal{T}_{pull} = [M_{1} \ \circled{c} \ M_{2} \ \circled{c} \ M_{3}] \\
  & \mathcal{T}_{f} = \mathcal{T}_{pull} + \mathcal{T}_{pull} \times \sigma(\mathcal{T}_{push})
  \label{fusion way}
\end{eqnarray}
$M_{1}, M_{2}, M_{3}$ represent upsampled mask embeddings generated from different scales of $\tfrac{1}{2}, \tfrac{1}{4}, \tfrac{1}{8}$ respectively, \circled{c} means the concatenation operation, $\mathcal{T}_{f}$ refers to decoded class tokens. Specifically, if the pulling force is stronger than the pushing force, then $\mathcal{T}_{f}$ will depict a picture of loose boundary representations, which are detrimental to the precise localization of inter-class boundaries. Under this circumstance, the pushing force needs to be strengthened. Otherwise, $\mathcal{T}_{f}$ will give a certain prediction for boundaries, which is not fitful for addressing boundary confusion, then we had better amplify the pulling force to introduce stronger boundary uncertainty. Therefore, a dynamic balance between these two adversarial forces tends to achieve an optimal state during the training process.

\section{Experimental Results}
\subsection{Datasets and Evaluation Metrics}
We conduct experiments on three public datasets and one private dataset to evaluate the segmentation performance for uncertain boundaries, including the pulmonary lobe dataset from LUNA16 \citep{tang2019automatic, setio2017validation, jacobs2014automatic}, the COVID-19 CT Lung and Infection Segmentation Dataset \citep{ma2021toward}, the Large Scale Vertebrae Segmentation Challenge (VerSe 2019) \citep{sekuboyina2021verse} and the left atrium and left atrial appendage dataset (LA/LAA) \citep{you2023semantic}.

\noindent\textbf{Pulmonary Lobe Dataset from LUNA16:} This dataset is collected from LUNA16, containing 51 CT scans. The reference annotation for each CT scan was manually delineated by radiologists using Chest Image Platform \url{https://chestimagingplatform.org/about}. And this dataset contains 6 segmentation classes (Class 0: the background, Class 1-2: the left upper and lower lobes, Class 3-5: the right upper, middle, and lower lobes). We employed 35 annotated CT scans for training, 6 cases for validation, and 10 cases for testing on the LUNA16 dataset. Here this dataset is noted as the clean lung lobe dataset for simplicity in our work.

\noindent\textbf{COVID-19 CT Lung and Infection Segmentation Dataset:} This dataset contains 20 COVID-19 CT scans. Left lung, right lung, and infections are labeled by two radiologists and verified by an experienced radiologist. Of all 20 CT scans, only 8 cases bear annotations for left and right pulmonary lobes. Besides, there are patchy shadows on lung lobes in CT scans, and some artifacts are located beside boundaries of two adjacent lobes. Considering the scarce data volumes, we mix this dataset with the Pulmonary Lobe Dataset from LUNA16 to evaluate models' segmentation performance. Specifically, 8 CT scans are randomly split as 4 cases for training and 4 cases for testing. Here this mixed dataset is noted as the fused pulmonary lobe dataset in our work, which contains 39 training cases, 6 validation cases, and 14 testing cases in total.

\begin{table*}[!t]
  \begin{center}
  \caption{Comparison with other models on fused lung lobes. (LU, LL, RU, RM, RL: left upper, left lower, right upper, right middle, and right lower lobes, L: lobes of the left lung, R: lobes of the right lung, Mean: the average evaluation metric of all lobes. Bold numbers: the best, Underlined numbers: the second best.)}
  \label{tab fused}
  \resizebox{0.95\textwidth}{!}{
  \begin{tabular}{ccccccccccccc}
  \hline  
 
  \multirow{2}*{Method} & \multicolumn{3}{c}{LUL} & \multicolumn{3}{c}{LLL}  & \multicolumn{3}{c}{L} & \multicolumn{3}{c}{Mean} \\  
  \cmidrule(r){2-4}  \cmidrule(r){5-7}  \cmidrule(r){8-10}  \cmidrule(r){11-13}
  & Dice & HD95 & ASSD & Dice & HD95 & ASSD & Dice & HD95 & ASSD & Dice & HD95 & ASSD \\
  
  \hline  
  3D UNet \citep{cciccek20163d} & 88.15 & 51.79 & 7.946
 & 86.44 & 41.92 & 6.344
 & 87.30 & 46.86 & 9.291
 & 69.37 & 48.69 & 9.291 \\
  VNet \citep{milletari2016v}  & 87.27 & 30.35 & 5.380
 & 87.41 & 44.90 & 6.429
 & 87.34 & 37.63 & 5.905
 & 67.54 & 41.55 & 8.931 \\
  ResUNet  \citep{diakogiannis2020resunet}   & 94.50 & 21.50 & 3.282
 & 93.74 & 13.08 & 2.706
 & 94.12 & 17.29 & 2.994
 & 71.24 & 30.56 & 6.210 \\
  TransUNet (3D)  \citep{chen2021transunet}   & 95.25 & 6.94 & 1.380
 & 94.39 & 24.66 & 3.298
 & 94.82 & 15.80 & 2.339
 & 70.73 & 31.64 & 5.773 \\
  TransBTS \citep{wang2021transbts}  & 87.64 & 30.87 & 5.192
 & 87.64 & 30.87 & 5.192
 & 86.52 & 36.89 & 5.890
 & 73.82 & 50.68 & 8.687 \\
  UNeXt  \citep{valanarasu2022unext}  & 94.75 & 12.17 & 2.260
 & 93.78 & 18.17 & 2.831
 & 94.26 & 15.17 & 2.546
 & 89.11 & 21.28 & 3.848  \\ 
  Swin UNETR \citep{tang2022self}  & 96.02 & 6.45 & 1.136
 & 95.50 & 6.53 & 1.172
 & 95.76 & 6.49 & 1.154
 & 92.82 & 7.78 & 1.622   \\
  3D UX-Net   \citep{lee20223d}  & 95.78 & 6.75 & 1.246
 & 95.33 & 7.84 & 1.407
 & 95.55 & 7.29 & 1.327
 & 92.48 & 10.36 & 1.917  \\
  nnUNet \citep{isensee2021nnu}  & \underline{96.73} & 5.78 & 1.032
 & \underline{96.30} & \underline{5.76} & 1.129
 & \underline{96.51} & \underline{5.77} & 1.081
 & 93.56 & 7.48 & 1.533 \\
  MedNeXt  \citep{roy2023mednext}  & \textbf{96.98} & \textbf{5.05} & \textbf{0.910}
 & \textbf{96.57} & \textbf{4.80} & \textbf{1.025}
 & \textbf{96.78} & \textbf{4.92} & \textbf{0.968}
 & \underline{93.75} & \underline{7.25} & \underline{1.517} \\
  Ours  & 96.56 & \underline{5.60} & \underline{0.973}
 & 96.26 & 6.19 & \underline{1.119}
 & 96.41 & 5.89 & \underline{1.046}
 & \textbf{94.52} & \textbf{6.56} & \textbf{1.317} \\
  \hline  
  \multirow{2}*{Method} & \multicolumn{3}{c}{RUL} & \multicolumn{3}{c}{RML}  & \multicolumn{3}{c}{RLL} & \multicolumn{3}{c}{R} \\  
  \cmidrule(r){2-4}  \cmidrule(r){5-7}  \cmidrule(r){8-10}  \cmidrule(r){11-13}
  & Dice & HD95 & ASSD & Dice & HD95 & ASSD & Dice & HD95 & ASSD & Dice & HD95 & ASSD \\
  \hline  

  3D UNet \citep{cciccek20163d}  & 64.48 & 65.67 & 13.119
 & 32.95 & 44.96 & 11.168
 & 74.83 & 39.09 & 7.876
 & 57.42 & 49.91 & 10.72 \\
  VNet \citep{milletari2016v}  & 46.08 & 53.83 & 14.979
 & 39.43 & 26.54 & 8.591
 & 77.49 & 52.11 & 9.274
 & 54.33 & 44.16 & 10.948  \\
  ResUNet  \citep{diakogiannis2020resunet}  & 57.34 & 40.02 & 8.555
 & 26.13 & 33.50 & 9.386
 & 84.50 & 44.73 & 7.120
 & 55.99 & 39.41 & 8.354  \\
  TransUNet (3D)  \citep{chen2021transunet}   & 37.51 & 46.74 & 10.066
 & 38.01 & 57.63 & 10.284
 & 88.48 & 22.25 & 3.837
 & 54.67 & 42.21 & 8.062 \\
  TransBTS \citep{wang2021transbts}   & 64.39 & 40.82 & 8.627
  & 53.08 & 29.48 & 7.581
 & 78.60 & 109.34 & 15.446
 & 65.36 & 59.88 & 10.551 \\
  UNeXt  \citep{valanarasu2022unext}  & 86.41 & 29.18 & 5.544
  & 79.11 & 19.15 & 4.422
 & 91.51 & 27.73 & 4.185
 & 85.68 & 25.35 & 4.717 \\
  Swin UNETR \citep{tang2022self}  & 93.69 & 8.27 & 1.602
 & 83.93 & \underline{12.06} & 3.065
 & 94.94 & \underline{5.60} & \underline{1.134}
 & 90.85 & 8.64 & 1.934 \\
  3D UX-Net   \citep{lee20223d}   & 93.68 & 8.04 & 1.687
 & 82.87 & 21.10 & 3.750
 & 94.73 & 8.06 & 1.492
 & 90.43 & 12.40 & 2.310 \\
  nnUNet \citep{isensee2021nnu}  & 93.96 & \underline{6.74} & \underline{1.396}
 & 85.49 & 12.93 & \underline{2.824}
 & 95.32 & 6.17 & 1.283
 & 91.59 & \underline{8.61} & \underline{1.834}  \\
  MedNeXt  \citep{roy2023mednext}  & \underline{93.97} & 7.13 & 1.434
 & \underline{85.66} & 12.70 & 2.836
 & \underline{95.59} & 6.55 & 1.380
 & \underline{91.74} & 8.79 & 1.883  \\
  Ours  & \textbf{95.01} & \textbf{5.83} & \textbf{1.180}
 & \textbf{88.76} & \textbf{9.78} & \textbf{2.191}
 & \textbf{96.58} & \textbf{4.33} & \textbf{0.941}
 & \textbf{93.45} & \textbf{6.65} & \textbf{1.438} \\

  \hline   
  \end{tabular}}
  \end{center}
\end{table*}

\noindent\textbf{VerSe 2019:} This CT dataset is composed of 80 training cases, 40 validation cases, and 40 testing cases. There are 26 segmentation classes, including label 0 for the background and label $1\mbox{-}25$ for 25 vertebrae. Of all 25 vertebrae, label $1\mbox{-}7$ represents cervical vertebrae, label $8\mbox{-}19$ for thoracic vertebrae and label $20\mbox{-}25$ for lumbar vertebrae. Different samples show different field of views (FOVs), which means they may have different kinds of vertebrae. 

\noindent\textbf{LA/LAA:} In detail, we privately collect 130 CT scans from 130 patients,
acquired by Siemens SOMATOM Force. Each CT volume consists of $256 \sim 528$ slices of $512 \times 512$ pixels, with a voxel space of $0.45\times0.49\times0.49$ $mm^{3}$. For annotation consistency, two clinicians individually annotated 65 CT cases, then one senior expert with over 20 years of experience corrected those annotations, especially the delineation of uncertain boundaries between the LA and LAA. The whole dataset is split into 70 training, 25 validation, and 35 testing cases. 

We evaluate PnPNet on testing 3D volumes. And we use both voxel overlap-based metrics and distance-based metrics. In terms of overlap-based metrics, we use the well-known Dice similarity coefficient (DSC) \citep{milletari2016v}, which ranges from 0\% (zero overlap) to 100\% (perfect overlap). In our results, we report the DSC averaged over all non-background
channels. The Hausdorff Distance (HD) \citep{karimi2019reducing} measures the quality of the segmentation by computing the maximum shortest distance between a point
from the prediction contour and a point from the target contour. And this metric can quantitatively reflect the segmentation performance on boundaries. Since
the Hausdorff Distance tends to be sensitive to outliers, we use a
more robust variant HD95 which considers the 95th percentile instead of the true maximum. Besides, we adopt the average symmetric surface distance (ASSD) \citep{yeghiazaryan2018family}, which measures the average distance between the surface of regions X and Y.

\begin{figure*}[!t]
\centerline{\includegraphics[width=0.85\linewidth]{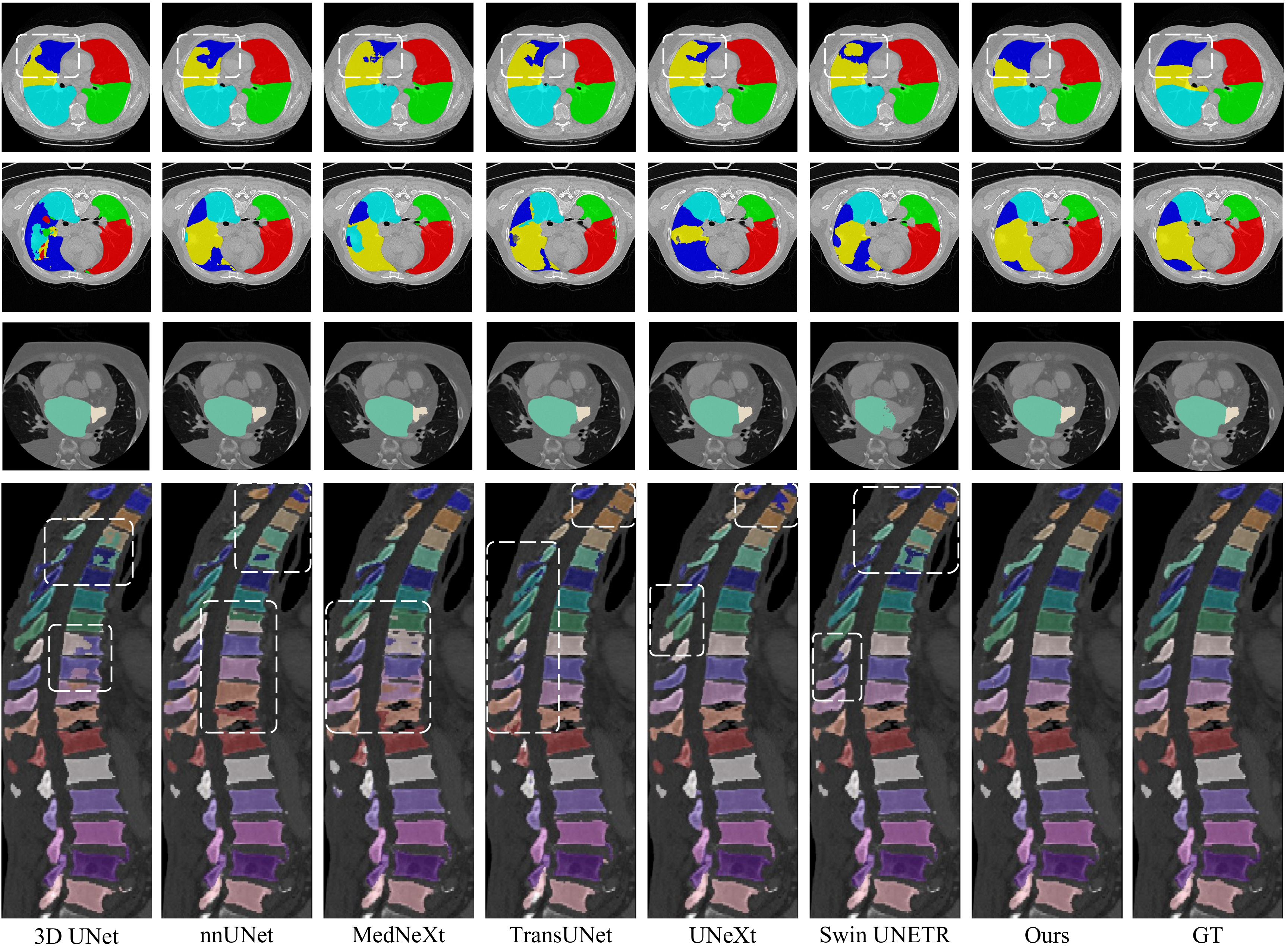}}
\caption{Qualitative visualizations. From the first row to the last row: the clean lung lobe, the fused lung lobe, LA/LAA, and VerSe 2019.}
\label{model visualizations}
\end{figure*}

\subsection{Implementation Details}
The proposed model is implemented with PyTorch 1.8.0 and trained on 2 NVIDIA Telsa V100, with a batch size of 2 in each GPU. For the clean pulmonary lobe dataset, all models are trained with the AdamW \citep{loshchilov2017decoupled} optimizer for 1500 epochs, with a warm-up cosine scheduler for the first 50 epochs. The initial learning rate is set as $5e\mbox{-}4$ with $1e\mbox{-}5$ weight decay. And the size of cropped patches is $16 \times 336 \times 448$. We do not utilize complicated data augmentations like previous works \citep{isensee2021nnu, zhou2021nnformer}. Instead, we adopt strategies of random rotation, random intensity shift and scale. We select ResUNet \citep{diakogiannis2020resunet} as the baseline model. For the fused pulmonary lobe dataset, the experimental setting is the same as that of the clean dataset. For the VerSe 2019 dataset, we train all models for 1000 epochs. All preprocessed cases are cropped with a patch size of $128 \times 160 \times 96$. Random rotation between [$-15^{\circ}, 15^{\circ}$] and random flipping along the XOZ or YOZ plane are employed for the data diversity. Here MedNeXt \citep{roy2023mednext} is chosen as the baseline model. For the LA/LAA dataset, models are trained for 1500 epochs and the patch size is set as $160 \times 160 \times 192$. Similarly, we augment the cardiac data with random rotation, random intensity shift and scale. The baseline model is set as MedNeXt. And for both VerSe 2019 and LA/LAA, the settings of the optimizer, the initial learning rate, and the learning rate scheduler remain the same as those for the clean pulmonary lobe dataset. For the choice of loss functions, we choose a weighted sum of Dice loss, cross-entropy loss, and class center loss for model training. And $\lambda$ is empirically set as $0.1$.
\begin{eqnarray}
  &  \mathcal L = \mathcal L_{Dice} + \mathcal L_{ce} + \lambda \times \mathcal L_{cc}
  \label{loss function}
\end{eqnarray}

\subsection{Results}
\subsubsection{Comparisons with other segmentation models}
We have evaluated our model's performance compared with classic CNNs \citep{cciccek20163d, isensee2021nnu, roy2023mednext}, recent Transformer-based \citep{chen2021transunet, wang2021transbts, tang2022self} and MLP-based \citep{valanarasu2022unext} segmentation models. And we provide quantitative and qualitative experimental results on the clean and fused lung lobe datasets, VerSe 2019, and LA/LAA CT scans.

\noindent\textbf{Evaluations on Clean Lung Lobe.} To fairly evaluate the performance of our model, especially on inter-class boundaries, we firstly conduct experiments on the clean lung lobe dataset. The quantitative results are illustrated in Table \ref{tab clean}. Our model achieves the best segmentation performance across the benchmark, with $95.35\%$, 6.11mm, and 1.171mm for the metrics of average Dice, HD95, and ASSD. Specifically, the recent state-of-the-art (SOTA) approach MedNeXt \citep{roy2023mednext} ranks the second best in the benchmark. And our model outperforms MedNeXt on quantitative evaluations for lobes of the left and right lungs. For lobes of the right lung, the proposed model outperforms MedNeXt with a $0.61\%$ Dice increase, 0.55mm HD95 decrease, and 0.187mm ASSD decrease.

\begin{table}[!t]
  \begin{center}
  \caption{Comparison with other models on VerSe 2019. (Cerv: Cervical vertebrae, Thor: Thoracic vertebrae, Lumb: Lumbar vertebrae, Mean: the average evaluation metric of all vertebrae. Bold numbers: the best, Underlined numbers: the second best.)}
  \label{tab verse}
    \resizebox{0.95\columnwidth}{!}{
  \begin{tabular}{ccccccccccc}
  \hline  
  \multirow{2}*{Method} & \multicolumn{4}{c}{Dice score (\%) $\uparrow$} & \multicolumn{4}{c}{$HD95$ (mm) $\downarrow$}  & \multirow{2}*{Params (M)} & \multirow{2}*{FLOPs (T)} \\  
  \cmidrule(r){2-5}  \cmidrule(r){6-9}
  & Cervical & Thoracic & Lumbar & Mean & Cervical & Thoracic & Lumbar & Mean \\
  \hline  
  3D UNet \citep{cciccek20163d}  & 83.10 & 78.37 & 70.88 & 81.28 & 3.26 & 6.27 & 8.50 & 5.80 & 16.49 & 0.521 \\
  VNet \citep{milletari2016v}  & 86.32 & 87.78 & 73.45 & 85.57 & 2.19 & 3.37 & 8.48 & 4.11 & 45.73 & 0.958 \\
  nnUNet \citep{isensee2021nnu}  & 87.81 & \underline{88.80} & \textbf{74.96} & 86.59 & 2.52 & 3.04 & \textbf{7.10} & 4.09 & 30.90  & 0.618 \\
  TransUNet (3D) \citep{chen2021transunet}  & 85.49 & 82.67 & 73.88 & 83.53  & 2.02 & 3.73 & 7.89 & 4.16 &  146.68 & 0.683 \\
  CoTr \citep{xie2021cotr}  & 81.48 & 79.68 & 68.83 & 80.59  & 3.92 & 9.88 & 14.34 & 9.05  & 48.53 &  0.508 \\
  UNeXt \citep{valanarasu2022unext}  & 77.00 & 86.73 & 71.06 & 83.36  & 3.44 & 2.97 & 9.47 & 4.43  & 4.02 & 0.012  \\
  Verteformer \citep{you2023verteformer} & 87.25 & 88.76 & 72.73 & 86.54 & 1.96 & \underline{2.87} & 8.21 & 3.66  & 330.65 & 0.336  \\
  Maskformer \citep{cheng2021per}  & 76.22 & 80.87 & 72.01 & 83.24   & 2.32 & 7.47 & 9.08 & 6.29   & 64.40 & 0.943 \\
  EG-Trans3DUNet \citep{you2022eg}  & 83.67 & 82.41 & 74.11 & 86.01  & 2.37 & 4.46 & 8.12 & 4.03  & 161.89 & 0.748 \\
  Swin UNETR \citep{tang2022self}  & \underline{89.30} & 81.43 & 73.36 & 83.46  & \textbf{1.85} & 5.90 & 8.81 & 5.75  & 62.19  & 0.732 \\
  MedNeXt \citep{roy2023mednext}  & 87.80 & 88.73 & 73.43 & \underline{87.29} & 2.32 & 2.88 & 8.06 & \underline{3.73} & 10.53 & 0.331 \\
  Ours  & \textbf{90.50} & \textbf{91.42} & \underline{74.14} & \textbf{88.71} & \underline{1.89} & \textbf{1.94} & \underline{7.67} & \textbf{3.04}  & 23.05 & 0.590 \\
  \hline  
  \end{tabular}}
  \end{center}
\end{table}

\noindent\textbf{Evaluations on Fused Lung Lobe.} To further assess the ability of our model to address boundary confusion, we carried out experiments on the fused lung lobe dataset. Lung lobe CT scans with COVID-19 infections bear uncertain boundaries with noise adjacent to them, which brings a huge challenge for the precise segmentation of inter-class boundaries. Besides, lung lobe CT datasets with COVID-19 infections show a different domain distribution from that of clean lung lobe datasets. Thus, it is more difficult to train a model to implement the segmentation task for the fused lung lobe. Table \ref{tab fused} illustrates the performance comparison between our model and other CNNs and Transformer-based models. We can discover that 3D UNet shows poor segmentation results for the fused lung lobe dataset, which results from the fact that the vanilla structure falls lack of strong representation abilities \citep{el2021high}. Detailedly, Figure \ref{model visualizations} reveals that 3D UNet cannot finely localize the boundaries between different parts of lobes. Compared with that, MedNeXt \citep{roy2023mednext} and nnUNet \citep{isensee2021nnu} can generate rich voxel-wise features to boost models' generalization abilities on this multi-domain dataset. Specifically, MedNeXt achieves a Dice score of $93.75\%$, a HD95 metric of 7.25mm, an ASSD metric of 1.517mm. As the recent SOTA model, MedNeXt outperforms other models on segmentation metrics for left lobes. However, for right lobes which have a more complicated structure especially on the inter-class boundaries, networks including nnUNet, 3D UX-Net, and Swin UNETR cannot well address the boundary confusion problem. By introducing the pushing and pulling branch into the baseline model, we achieve a $0.77\%$ Dice increase, 0.69mm HD95 decrease, and 0.200mm ASSD decrease compared with MedNeXt. And it is worth mentioning that PnPNet improves a lot on the metrics for right lobes, with a $1.71\%$ Dice increase, 2.14mm HD95 decrease, and 0.445mm ASSD decrease. For the right middle lobe with a larger shape variance, our model outperforms nnUNet with a $3.27\%$ Dice increase, 3.15mm HD95 decrease, and 0.645mm ASSD decrease.

\begin{table}[!t]
  \begin{center}
  \caption{Comparison with other models on LA/LAA. (LA: Left Atrium, LAA: Left Atrium Appendage, Mean: the average evaluation metric of the LA and LAA. Bold numbers: the best, Underlined numbers: the second best.)}
  \label{tab LA}
  \resizebox{0.75\columnwidth}{!}{
  \begin{tabular}{ccccccccc}  
  \hline  
  \multirow{2}*{Model} &  \multicolumn{3}{c}{Dice score (\%) $\uparrow$} & \multicolumn{3}{c}{$HD95$ (mm) $\downarrow$} & \multirow{2}*{\scriptsize{Params(M)}} & \multirow{2}*{\scriptsize{FLOPs(T)}}  \\  
  \cmidrule(r){2-4}    \cmidrule(r){5-7}
  & LAA & LA & Mean & LAA & LA & Mean  & & \\
  \hline  
  3D UNet \citep{cciccek20163d}  & 83.88 & 94.40 & 89.14 & 3.97 & 5.48 & 4.72 & 16.47 & 1.206 \\  
  ResUNet \citep{diakogiannis2020resunet}  & 83.07 & 95.65 & 89.36 & 3.94 & 3.49 & 3.71 & 27.19 & 0.837 \\
  nnUNet \citep{isensee2021nnu} & \underline{84.06} & 95.57 & 89.81 & 3.95 & 3.32 & 3.63 & 30.79 & 1.252  \\
  3D UX-Net \citep{lee20223d}  & 83.52 & 95.59 & 89.55 & 4.05 & 3.44 & 3.74  & 53.01 & 3.510 \\
  MedNeXt \citep{roy2023mednext}  & 84.00 & \underline{95.66} & \underline{89.83} & 3.82 & \underline{3.28} & \underline{3.55}  & 10.52 & 0.433 \\
UNeXt \citep{valanarasu2022unext}  & 83.03 & 95.16 & 89.09 & 3.98 & 3.52 & 3.75  & 4.02 & 0.035 \\
  TransUNet (3D) \citep{chen2021transunet}  & 83.46 & 95.28 & 89.37 & 4.17 & 4.41 & 4.29  & 321.36 & 2.398 \\
  Swin UNETR \citep{tang2022self} & 83.32 & 95.53 & 89.43 & \underline{3.81} & 3.29 & \underline{3.55}   & 63.96 & 1.827 \\
  TransBTS \citep{wang2021transbts}  & 70.17 & 93.21 & 81.69 & 6.05 & 7.53 & 6.79  & 35.41 & 0.614 \\
  Ours  & \textbf{84.51} & \textbf{95.73} & \textbf{90.12} & \textbf{3.78} & \textbf{3.14} & \textbf{3.46}  & 22.86  &  0.741 \\
  \hline 
  \end{tabular}}
  \end{center}
\end{table}

\noindent\textbf{Evaluations on VerSe 2019.} We also evaluate the proposed PnPNet on the hidden test dataset of VerSe 2019. As shown in Figure \ref{model visualizations}, deep segmentation models suffer from the challenge that adjacent vertebrae are not precisely separated by two boundary interfaces. As a result, there is a segmentation inconsistency inside vertebrae. In our model, the pulling branch focuses on squeezing each vertebral region, and is aimed at addressing the challenge of segmentation inconsistency. Besides, the pushing branch helps to precisely localize the boundary region between two adjoining vertebrae. Table \ref{tab LA} shows that PnPNet achieves the best performance on the cervical, thoracic, and average metrics, with a $2.70\%$, $2.69\%$, and $1.42\%$ Dice increase respectively. Besides, different CT scans have different field-of-views (FoVs), which makes it difficult for models to identify the last several lumbar vertebrae (Label $20\mbox{-}25$). And nnUNet is superior to other models on the ability to localize and segment lumbar vertebrae. Figure \ref{model visualizations} demonstrates that PnPNet gives the best visualization results among all models, in which 3D UNet and UNeXt exhibit inconsistent predictions, TransUNet even gives continuous wrong predictions for the label of vertebrae.

\noindent\textbf{Evaluations on LA/LAA.} To further validate the effectiveness and robustness of PnPNet, we conduct experiments on the LA/LAA dataset, which is deficient in the uniform standard for annotations of uncertain boundaries. According to Table \ref{tab LA}, our model achieves the highest average Dice score of $90.12\%$ and the lowest average HD95 value of 3.46mm. And the standard deviation of these two metrics is also lower than that of other models, which reveals that PnPNet can improve the segmentation results of the whole dataset to some degree. Here the structure of the left atrial appendage bears various anatomical shapes. Thus, the delineation of the LAA is challenging. Our model surpasses MedNeXt on the Dice score of the LAA with a $0.51\%$ increase and a smaller standard deviation value.

\subsubsection{Comparisons with other modules on refining boundary confusion}
To evaluate the efficacy of our proposed SDM compared with other modules on refining the uncertain boundaries, we conduct experiments on the clean and fused lung lobe datasets by adding these modules to the baseline model. These modules are devised to enhance feature representations of boundaries, highlighting regions with boundary confusion. We 
select attention modules including the boundary preserving block (BPB) \citep{lee2020structure}, the adaptive fusion module (AFM) \citep{xie2022uncertainty}, the recurrent edge correction (REC) \citep{xie2022uncertainty}, the topological interaction module (TIM) \citep{gupta2022learning}, and the boundary enhancement module (BEM) \citep{lin2023rethinking}, the weighted attention module (WAM) \citep{wang2023xbound}.

As shown in Table \ref{modules on refining boundaries}, introducing the proposed SDM will largely promote the segmentation performance of the baseline model. For the clean lung lobe dataset, SDM brings a $1.10\%$ Dice increase and 0.371mm ASSD decrease. While for the fused lung lobe, the significant improvement achieves a $22.95\%$ Dice increase and a 4.801mm ASSD decrease. Besides, compared with other modules, our proposed SDM can better improve the baseline model on the evaluation metric of the ASSD value, which demonstrates that SDM manifests a stronger ability to enhance the boundary representations.

\begin{table}[!t]
  \begin{center}
  \caption{Comparisons with other implicit enhancement modules on refining uncertain boundaries for clean and fused lung lobes. (L: lobes of Left lung, R: lobes of Right lung, Mean: the average evaluation metric.)}
  \label{modules on refining boundaries}
  \resizebox{0.95\columnwidth}{!}{
  \begin{tabular}{cccccccccc}
  \hline  
  \multicolumn{10}{c}{(a) Clean Lung Lobes} \\
    \hline    
    \multirow{2}*{Method} & \multicolumn{3}{c}{L} & \multicolumn{3}{c}{R} & \multicolumn{3}{c}{Mean} \\  
  \cmidrule(r){2-4}  \cmidrule(r){5-7}  \cmidrule(r){8-10} 
  & Dice$\uparrow$ & HD95$\downarrow$ & ASSD$\downarrow$ & Dice$\uparrow$ & HD95$\downarrow$ & ASSD$\downarrow$ & Dice$\uparrow$ & HD95$\downarrow$ & ASSD$\downarrow$ \\
  \hline  
  Baseline  & 96.66 & 6.38 & 1.214
 & 92.40 & 8.17 & 1.980
 & 94.11 & 7.45 & 1.674  \\
  + BPB \citep{lee2020structure}
  & 96.45 & 5.76 & 1.161
 & 93.29 & \textbf{7.97} & 1.733
 & 94.55 & 7.05 & 1.504 \\
 + AFM \citep{xie2022uncertainty}
 & 97.63 & 4.41 & 0.857
 & 92.40 & 8.28 & 1.742
 & 94.49 & 6.74 & 1.388 \\
 + REC \citep{xie2022uncertainty}
 & 97.20 & 4.90 & 0.977
 & 93.15 & 8.52 & 1.792
 & 94.77 & 7.07 & 1.466 \\
 + TIM \citep{gupta2022learning}
 & 96.92 & 5.55 & 1.004
 & 93.48 & 8.13 & 1.747
 & 94.86 & 7.10 & 1.450 \\
 + BEM \citep{lin2023rethinking}
 & 97.55 & 3.98 & 0.806
 & 93.12 & 8.37 & 1.724
 & 94.89 & 6.61 & 1.357 \\
 + WAM \citep{wang2023xbound}
 & 97.46  &  3.94 & 0.769
 & 93.28  &  9.23 & 1.824
 & 94.95  & 7.11  & 1.402 \\
 + SDM (ours)
  & \textbf{97.83} & \textbf{3.41} & \textbf{0.760}
 & \textbf{93.47} & 8.37 & \textbf{1.666}
 & \textbf{95.21} & \textbf{6.38} & \textbf{1.303} \\
  \hline  
    \multicolumn{10}{c}{(b) Fused Lung Lobes} \\
    \hline  
      Baseline  & 94.12 & 17.29 & 2.994
 & 55.99 & 39.41 & 8.354
 & 71.24 & 30.56 & 6.210 \\
  + BPB \citep{lee2020structure}
  & 96.46 & 9.26 & 1.908
 & 84.98 & 33.93 & 5.529
 & 89.57 & 24.06 & 4.081 \\
 + AFM \citep{xie2022uncertainty}
 & 96.88 & 5.38 & 1.088
 & 92.26 & 8.49 & 1.744
 & 94.11 & 7.25 & 1.481 \\
 + REC \citep{xie2022uncertainty}
 & 87.24 & 38.40 & 5.941
 & 57.57 & 43.90 & 9.350
 & 69.43 & 41.70 & 7.987 \\
 + TIM \citep{gupta2022learning}
 & 96.74 & 6.55 & 1.291
 & 92.25 & 8.64 & 1.783
 & 94.04 & 7.81 & 1.587 \\
 + BEM \citep{lin2023rethinking}
 & 96.29 & 6.54 & 1.295
 & 91.99 & 9.38 & 1.896
 & 93.71 & 8.24 & 1.656 \\
  + WAM \citep{wang2023xbound}
 & 96.01 & 6.53 & 1.193
 & 91.70  &  10.02 & 1.904
 & 93.43  &  8.62  & 1.619 \\
 + SDM (ours)
  & \textbf{96.95} & \textbf{5.27} & \textbf{0.995}
 & \textbf{92.36} & \textbf{7.94} & \textbf{1.685}
 & \textbf{94.19} & \textbf{6.87} & \textbf{1.409} \\
 \hline
  \end{tabular}}
  \end{center}
\end{table}

\begin{figure*}[!t]
\centerline{\includegraphics[width=1.0\linewidth]{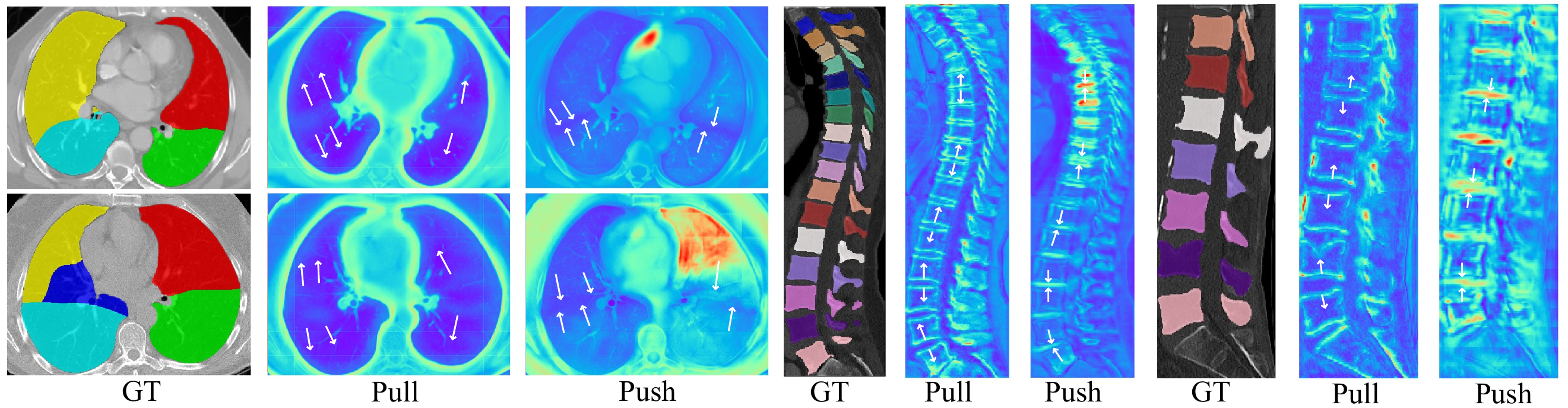}}
\caption{Visualizations of pulling and pushing tokens on pulmonary lobes and VerSe 2019. Pull: the pulling tokens from the pulling branch. Push: the pushing tokens from the pushing branch.}
\label{pull-push-token}
\end{figure*}

\begin{figure}
  \centering
  \includegraphics[width=0.60\linewidth]{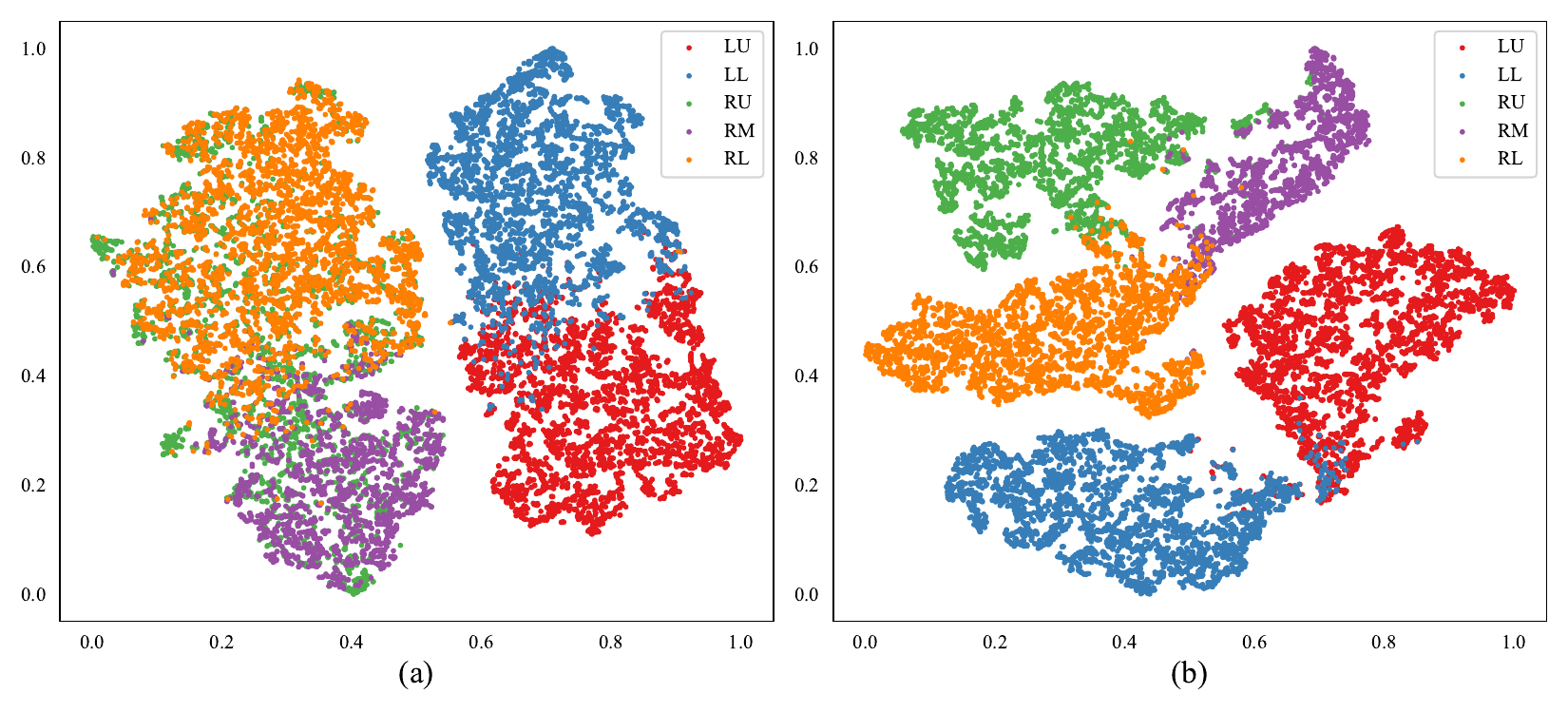}
  \hspace{1in}
  \includegraphics[width=0.60\linewidth]{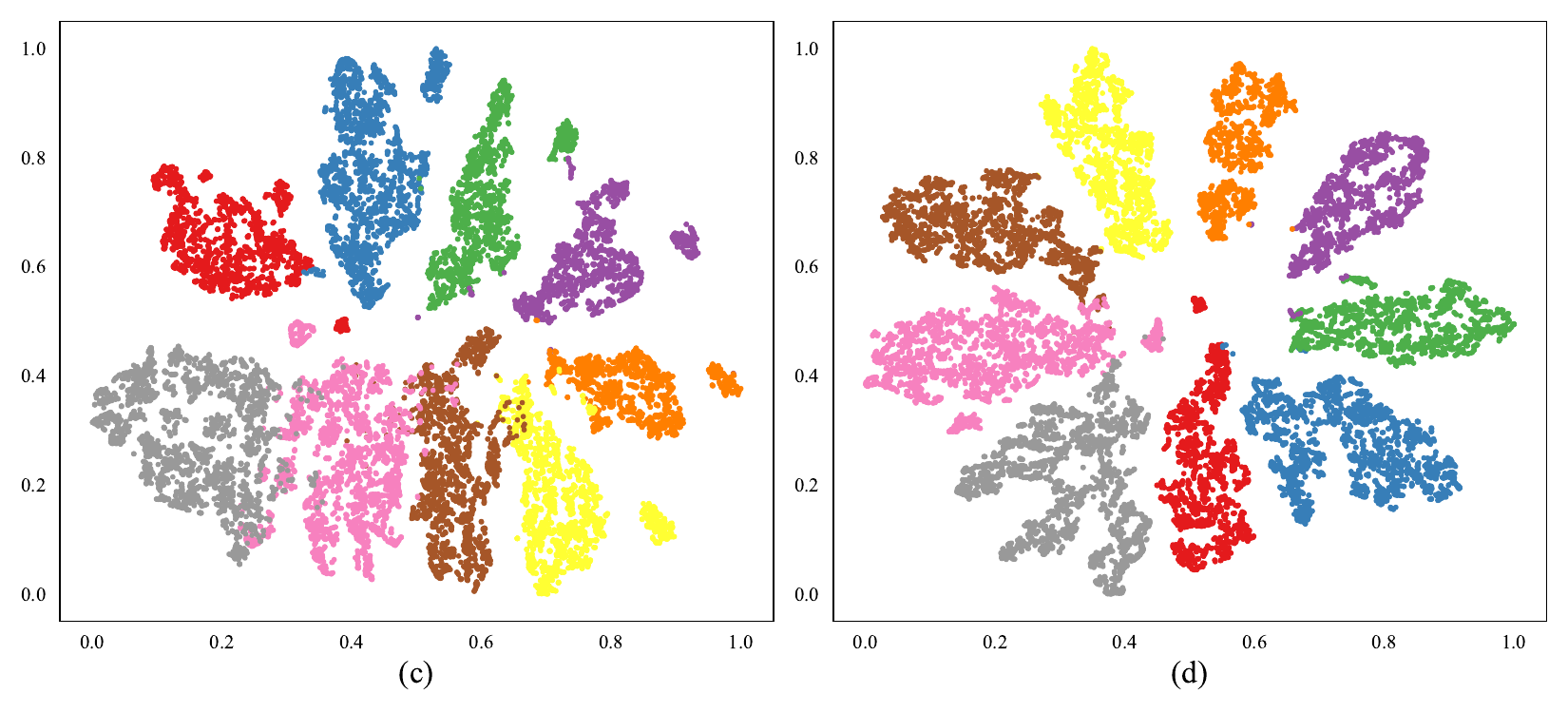}
  \caption{The T-SNE visualizations of the decoded feature embeddings. (a) Fused lung lobe with the baseline model. (b) Fused lung lobe with the baseline model integrated with SDM. (c) A vertebrae case with the baseline model. (d) A vertebrae case with the baseline model combined with CCM.}
  \label{t-sne}
\end{figure}

\subsection{Ablation Analysis}
\subsubsection{Ablation on Key Components}
\begin{figure}[!t]
\centerline{\includegraphics[width=0.55\linewidth]{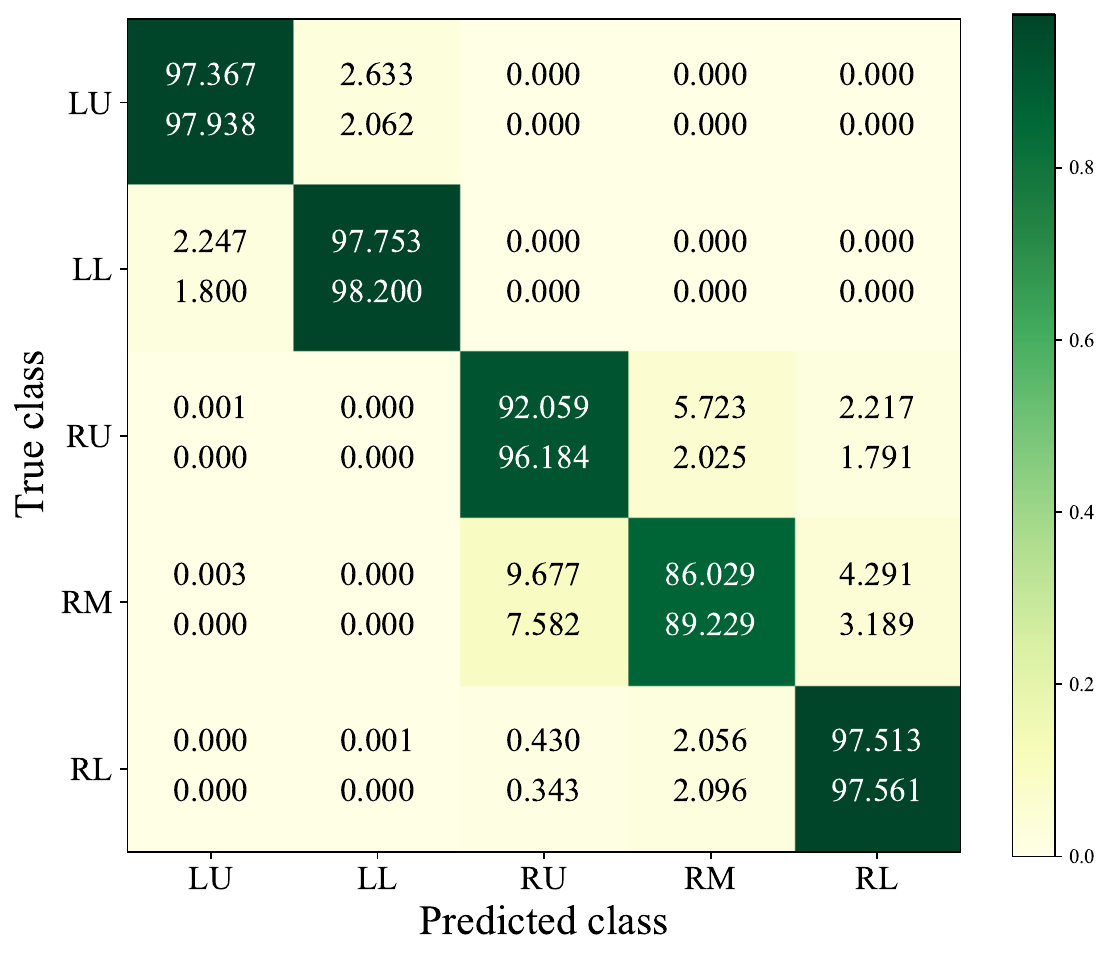}}
\caption{Confusion matrix corresponding to clean lung lobes between the baseline mode (the upper row) and the baseline + SDM mode (the lower row).}
\label{confusion_matrix}
\end{figure}

Firstly, we investigate the effectiveness of two proposed components (SDM and CCM) on lobe and vertebrae data. Employing the proposed SDM makes our PnPNet more powerful for modeling complicated boundary regions. And SDM attempts to enhance boundary feature representations by introducing pushing forces based on diffusion theory. For VerSe 2019, incorporating SDM into the baseline model will boost the segmentation performance on the cervical, thoracic, and lumbar vertebrae, with $3.08\%$ and $2.16\%$ Dice increases on cervical and thoracic vertebrae. For the clean lung lobe, we calculate the confusion matrix to measure the segmentation performance, especially for right lobes. As depicted in Figure \ref{confusion_matrix}, some voxels in the right middle lobe are wrongly classified as categories of right upper and lower lobes, and parts of the right upper lobe are prone to misclassification as the right middle lobe. And introducing SDM can substantially improve the problem above. For the fused lung lobe, we provide T-SNE visualizations of the decoded feature in Figure \ref{t-sne}. We can figure out that class-wise features are better divided into five different semantic classes after the baseline model is integrated with SDM. And this phenomenon indicates that SDM can enlarge the inter-class distance, then discriminate the inter-class boundaries between different anatomies.

\begin{table}[!t]
  \begin{center}
  \caption{Ablation studies for key components on clean and fused lung lobes, VerSe 2019. (L: lobes of the left lung, R: lobes of the right lung, RM: the right middle lobe, Cerv: Cervical vertebrae, Thor: Thoracic vertebrae, Lumb: Lumbar vertebrae, Mean: the average evaluation metric.)}
  \label{ablation for key components}
  \resizebox{0.70\columnwidth}{!}{
  \begin{tabular}{ccccccccc}
  \hline
    \multicolumn{9}{c}{(a) Clean Lung Lobes} \\
      \hline
  \multirow{2}*{Settings} & \multicolumn{2}{c}{L} & \multicolumn{2}{c}{RM} & \multicolumn{2}{c}{R} & \multicolumn{2}{c}{Mean} \\  
  \cmidrule(r){2-3}  \cmidrule(r){4-5}  \cmidrule(r){6-7}   \cmidrule(r){8-9}
  & Dice$\uparrow$ & ASSD$\downarrow$ & Dice$\uparrow$ & ASSD$\downarrow$ & Dice$\uparrow$ & ASSD$\downarrow$ & Dice$\uparrow$ & ASSD$\downarrow$ \\
    \hline
  Baseline  & 96.66 & 1.214
 & 86.85 & 3.219
 & 92.40 & 1.980
 & 94.11 & 1.674  \\
  + SDM  & \textbf{97.83} & \textbf{0.760}
  & 88.40 & 2.599
 & 93.47 & 1.666
 & 95.21 & 1.303 \\
  + CCM  & 97.36 & 0.860
  & 87.95 & 2.739
 & 93.49 & 1.463
 & 94.93 & 1.366 \\
  + Both  & 97.53 & 0.808
  & \textbf{88.99} & \textbf{2.519}
 & \textbf{93.90} & \textbf{1.413}
 & \textbf{95.35} & \textbf{1.171} \\

  \hline  
    
  \multicolumn{9}{c}{(b) Fused Lung Lobes} \\

  \hline  
  Baseline  & 94.12 & 2.994
 & 26.13 & 9.386
 & 55.99 & 8.354
 & 71.24 & 6.210  \\
  + SDM  & \textbf{96.95} & \textbf{0.995}
  & 87.98 & 3.085
 & 92.36 & 1.685
 & 94.19 & 1.409 \\
  + CCM & 96.23 & 1.106
  & 87.08 & 2.681
 & 92.43 & 1.747
 & 93.95 & 1.491 \\
  + Both  & 96.41 & 1.046
  & \textbf{88.76} & \textbf{2.191}
 & \textbf{93.45} & \textbf{1.438}
 & \textbf{94.52} & \textbf{1.317} \\
  \hline  
  \multicolumn{9}{c}{(c) VerSe 19} \\
      \hline   
  \multirow{2}*{Settings} & \multicolumn{2}{c}{Cerv} & \multicolumn{2}{c}{Thor} & \multicolumn{2}{c}{Lumb}  & \multicolumn{2}{c}{Mean} \\  
  \cmidrule(r){2-3} \cmidrule(r){4-5} \cmidrule(r){6-7} \cmidrule(r){8-9}
  & Dice$\uparrow$ & HD95$\downarrow$ & Dice$\uparrow$ & HD95$\downarrow$ & Dice$\uparrow$ & HD95$\downarrow$ & Dice$\uparrow$ & HD95$\downarrow$ \\

  \hline  
  Baseline  & 87.80 & 2.32 & 88.73 & 2.88 & 73.43 & 8.06 & 87.29 & 3.73  \\
  + SDM   & \textbf{90.88} & \textbf{1.73}  & 90.89  & 2.06 & 73.77 & 7.75 & 88.14  & 3.14  \\
  + CCM  & 88.40 & 1.97  & 89.31  & 2.80  & 73.72 & 7.98 & 87.94 & 3.28  \\
  + Both   & 90.50 & 1.89  & \textbf{91.42}  & \textbf{1.94} & \textbf{74.14} & \textbf{7.67} & \textbf{88.71} & \textbf{3.04}  \\
  \hline  
  \end{tabular}}
  \end{center}
\end{table}

Then we make a thorough inquiry on the efficacy of CCM. Incorporating the pulling branch into the baseline model leads to considerable performance improvements for lung lobe datasets. Specifically, for the fused lung lobe, the structure of CCM brings $2.11\%$ and $36.44\%$ Dice increases on the left and right lobes. While for the vertebrae data, CCM serves as a pulling force to stretch the intersected boundary region, then squeezes each vertebra to improve the segmentation consistency. As illustrated by Figure \ref{t-sne}, the class-wise features can be better divided into different classes. On the one hand, the intra-class distance is largely reduced, which means that CCM can tighten data distributions of each semantic class. On the other hand, inter-class distance has been expanded, which indicates that inter-class boundaries between anatomies can be more precisely segmented.

\begin{figure}[!t]
\centerline{\includegraphics[width=0.95\linewidth]{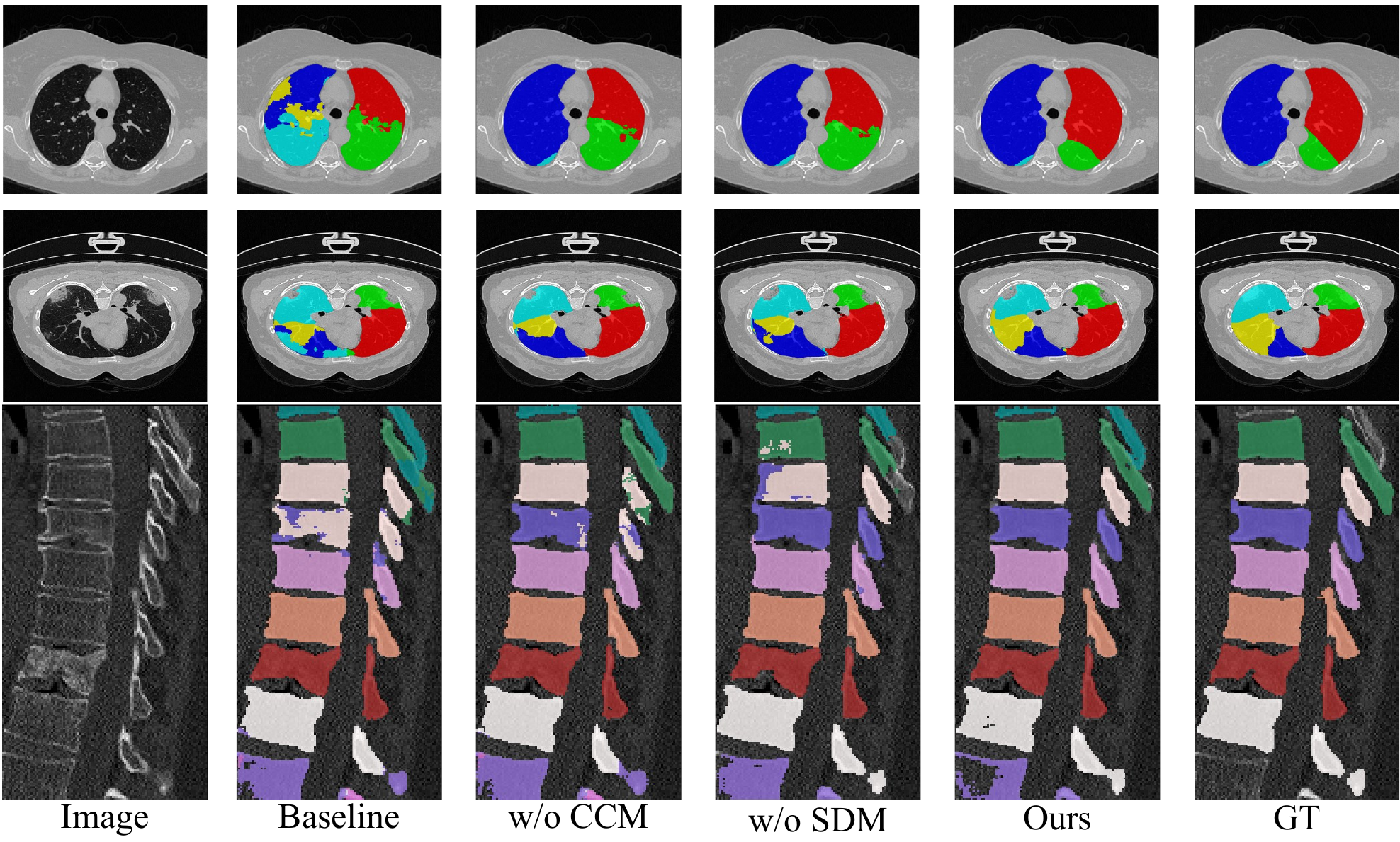}}
\caption{Ablation visualizations for key components on clean and fused pulmonary lobes, VerSe 2019.}
\label{key component ablation}
\end{figure}

\begin{figure*}[!t]
\centerline{\includegraphics[width=0.95\linewidth]{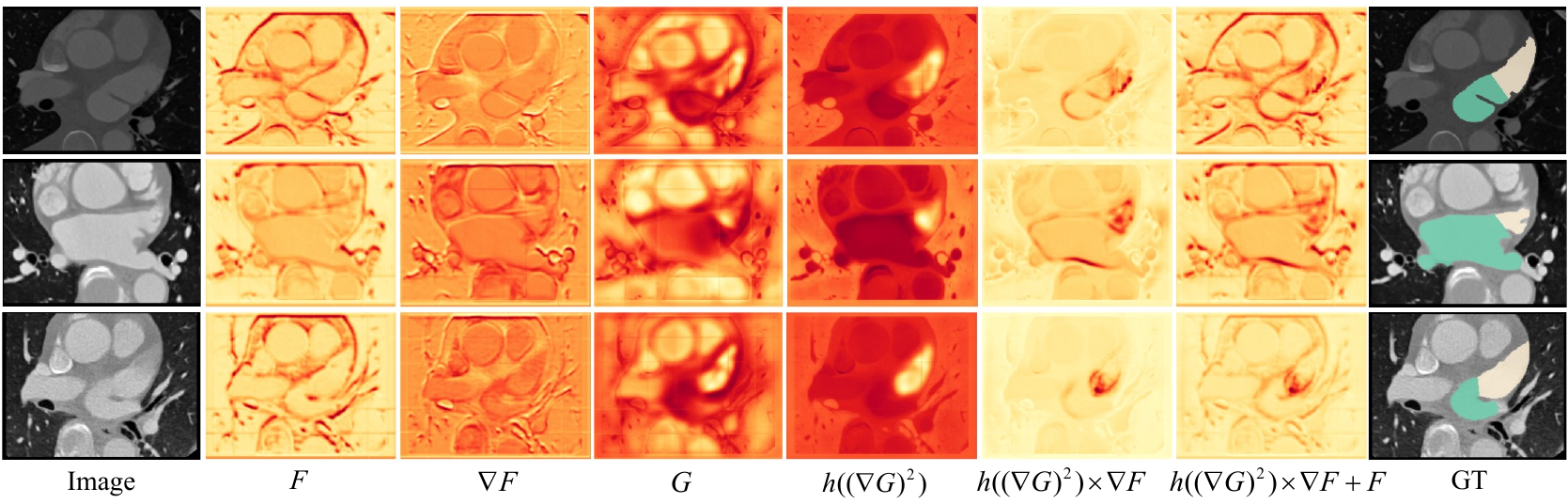}}
\caption{The detailed feature visualizations of key components in the structure of SDM when plugged into MedNeXt. And the evaluation is based on the LA and LAA. F: Raw skipped features. $\nabla F$: Raw differential features. G: Deep features from the precedent decoder layer. $h((\nabla G)^{2})$: Semantic guidance. $h((\nabla G)^{2}) \times \nabla F$: Highlighted boundary features. $h((\nabla G)^{2}) \times \nabla F + F$: Enhanced skipped features.}
\label{SDM feature}
\end{figure*}

By combining these two modules, the segmentation performance on boundary predictions is further boosted due to adversarial forces. As illustrated in Figure \ref{pull-push-token}, pulling tokens tend to enlarge the inter-class boundary region to amplify the boundary uncertainty, while pushing tokens try to squeeze it. Thus, boundaries between different anatomies are more precisely segmented after reaching the dynamic equilibrium state. Here we pay more attention to the evaluation metric on boundary surfaces. The ASSD value has decreased from 1.674mm to 1.171mm for the clean lung lobe, from 6.210mm to 1.317mm for the fused lung lobe, and from 3.73mm to 3.04mm for the vertebrae. This improvement reveals that adversarial pulling and pushing forces help to more precisely localize uncertain boundaries. However, after introducing CCM to the baseline model with SDM, the segmentation performance for cervical vertebrae declines slightly while thoracic and lumbar vertebrae are finely
segmented, which might result from the biased ratio between three kinds of vertebrae \citep{sekuboyina2021verse}. And the same situation comes to the segmentation performance of left lobes. We argue that CCM is essentially effective in clustering five lobes, while class distributions of right lobes are different from those of left lobes. And neural networks exert the boundary shape constraint of right lobes to left lobes due to the class imbalance. Furthermore, Figure \ref{key component ablation} exhibits the ablation visualizations for key components with a vertebral CT case. Adding SDM and CCM achieves a finer and more consistent prediction, especially for boundaries.

\subsubsection{Ablation on the Structure of SDM}
In addition, we carry out a supplementary analysis on the structure of SDM. SDM inherently enhances raw skipped features with abundant boundary information. Thus, we emphasize three core ingredients, which are raw features, the EID kernel, and semantic guidance in respective. As revealed in Table \ref{structure ablation}, we evaluate the structural function of each component with quantitative results on VerSe 2019. Specifically, removing original features and the semantic guidance will both degrade the segmentation performance on cervical, thoracic, and lumbar vertebrae. Besides, qualitative visualization results are illustrated in Figure \ref{SDM structure}. Original differential features $\nabla F$, containing abundant boundary information, are attained from raw skipped features $F$ after being processed with the learnable difference kernel $K_{o}$. However, original differential features involve irrelevant activations, and that is why semantic guidance is introduced for refinement. And semantic information is generated from the differential operation of deeper features $G$ from the precedent decoder layer. Here we can observe that semantic guidance highlights inter-class boundaries and suppresses the region of hilus pulmonis. Then after continuous element-wise addition and element-wise multiplication operations, raw skipped features are transformed into enhanced features with richer boundary information, which is beneficial to the precise segmentation of uncertain boundaries. More detailed visualization results on the LA and LAA are illustrated in Figure \ref{SDM feature}. Furthermore, to avoid the exception that all values in kernels are negative, we proposed the explicit-implicit differential kernel to enhance the representation of boundary information. As shown in Table \ref{structure ablation}, improving the original kernel will bring a $2.02\%$ and $0.60\%$ Dice increases on the thoracic and average metrics.

\begin{table}[!t]
  \begin{center}
  \caption{Ablation study on the structure of SDM and the pulling branch for VerSe 2019 and fused lung lobes respectively. (L: lobes of the left lung, R: lobes of the right lung, Cerv: Cervical vertebrae, Thor: Thoracic vertebrae, Lumb: Lumbar vertebrae, Mean: the average evaluation metric.)}
  \label{structure ablation}
  \resizebox{0.80\columnwidth}{!}{
  \begin{tabular}{ccccccccc}
  \hline  
  \multicolumn{9}{c}{(a) Structural ablations for SDM on VerSe19} \\
  \hline  
  \multirow{2}*{Settings} & \multicolumn{2}{c}{Cerv} & \multicolumn{2}{c}{Thor} & \multicolumn{2}{c}{Lumb}  & \multicolumn{2}{c}{Mean} \\  
  \cmidrule(r){2-3} \cmidrule(r){4-5} \cmidrule(r){6-7} \cmidrule(r){8-9}
  & Dice$\uparrow$ & HD95$\downarrow$ & Dice$\uparrow$ & HD95$\downarrow$ & Dice$\uparrow$ & HD95$\downarrow$ & Dice$\uparrow$ & HD95$\downarrow$ \\ 
  \hline  
    + SDM & \textbf{90.88} & \textbf{1.73}  & \textbf{90.89} & \textbf{2.06} & \textbf{73.77} & \textbf{7.75} & \textbf{88.14} & \textbf{3.14}  \\
    original kernel & 90.77 & 1.74 & 88.87 & 2.56 & 73.55 & 7.79 & 87.54 & 3.49 \\
    w/o original features & 87.35 & 2.59 & 88.05 & 2.75 & 73.49 & 8.08 & 87.37 & 3.48 \\
    w/o semantic guidance & 89.76 & 2.15 & 89.94 & 2.28 & 72.99 & 8.12 & 87.61 & 3.64 \\
  \hline  
    \multicolumn{9}{c}{(b) Structural ablations for CCM on fused lung lobes} \\
  \hline  
  \multirow{2}*{Settings} & \multicolumn{2}{c}{L} & \multicolumn{2}{c}{R-M} & \multicolumn{2}{c}{R}  & \multicolumn{2}{c}{Mean} \\  
  \cmidrule(r){2-3} \cmidrule(r){4-5} \cmidrule(r){6-7} \cmidrule(r){8-9}
  & Dice$\uparrow$ & HD95$\downarrow$ & Dice$\uparrow$ & HD95$\downarrow$ & Dice$\uparrow$ & HD95$\downarrow$ & Dice$\uparrow$ & HD95$\downarrow$ \\
 
  \hline  
    + CCM & 96.23 & 6.02  & \textbf{86.94} & \textbf{11.54} & \textbf{92.43} & \textbf{8.19} & \textbf{93.95} & \textbf{7.32} \\
    w/o supervision & \textbf{96.40} & \textbf{5.85} & 84.83 & 12.16 & 91.25 & 8.64 & 93.31 & 7.52 \\
    w/o center atlas & 95.85 & 7.13 & 86.24 & 12.47 & 91.89 & 8.92 & 93.47 & 8.20 \\
    w/o both & 95.43 & 7.91 & 84.73 & 12.62 & 91.22 & 9.27 & 92.91 & 8.73 \\
  \hline  
  \end{tabular}}
  \end{center}
\end{table}

\begin{figure}[!t]
  \centering
  \begin{minipage}{.40\linewidth}
    \centering
    \includegraphics[width=0.95\linewidth]{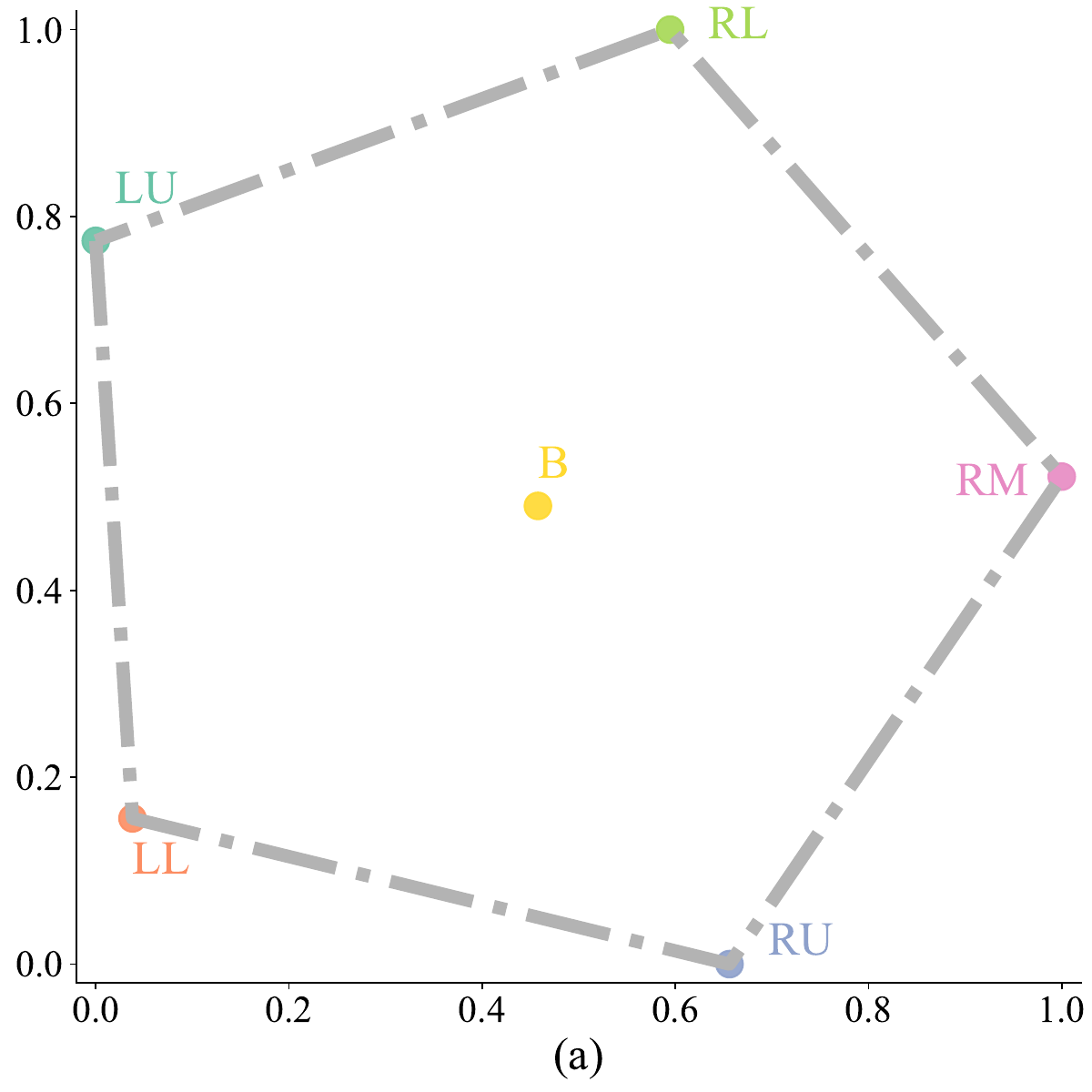}
  \end{minipage}%
  \begin{minipage}{.40\linewidth}
    \centering
    \includegraphics[width=0.95\linewidth]{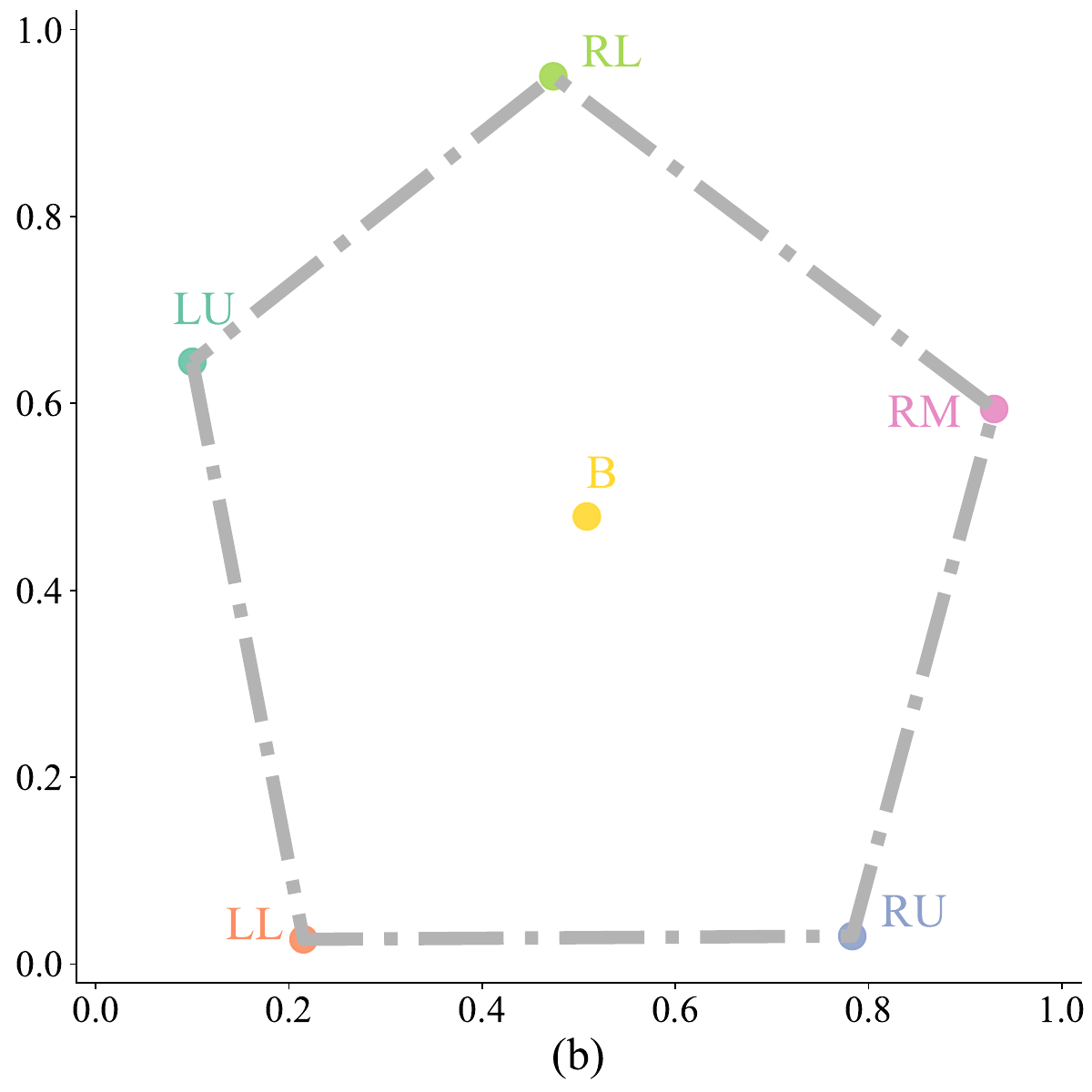}
  \end{minipage}
    \caption{Visualizations of class centers after T-SNE. (a) Clustered class centers from the center atlas in the baseline model incorporated with CCM. (b) Initial class centers in the baseline model. LU, LL, RU, RM, RL: Left Upper, Left Lower, Right Upper, Right Middle, and Right Lower Lobes. B: Background.}
  \label{class center}
\end{figure}

\begin{figure*}[!t]
\centerline{\includegraphics[width=0.95\linewidth]{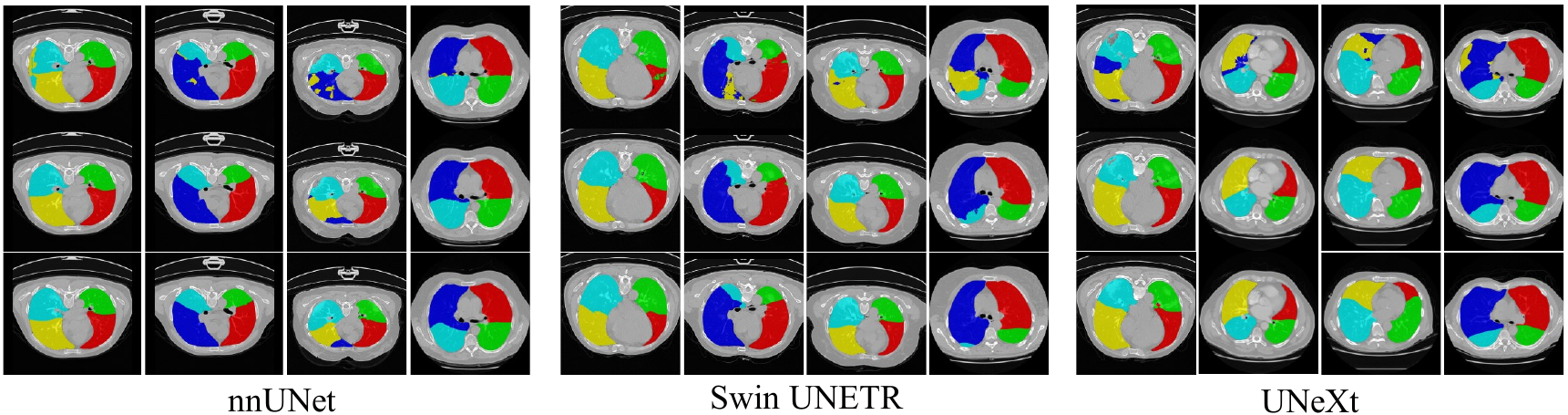}}
\caption{Qualitative visualizations on fused lung lobes when SDM and CCM are plugged into other segmentation backbones. Here we choose nnUNet, Swin UNETR, and UNeXt from the family of CNNs, Transformer-based and MLP-based models respectively. The first row: predictions from the baseline model. The second row: predictions from the baseline model plus SDM and CCM. The third row: GTs.}
\label{ccm-sdm plug-and-play visualizations}
\end{figure*}

\begin{table}[!t]
  \begin{center}
  \caption{Ablation study on the plug-and-play property of SDM for clean lung lobes , and SDM plus CCM for fused lung lobes. (L: lobes of the left lung, R: lobes of the right lung, Mean: the average evaluation metric.)}
  \label{tab sdm plug-and-play}
  \resizebox{0.80\columnwidth}{!}{
  \begin{tabular}{cccccccccc}
  \hline
    \multicolumn{10}{c}{(a) SDM for clean lung lobes} \\
      \hline  
    \multirow{2}*{Method} & \multicolumn{3}{c}{L} & \multicolumn{3}{c}{R} & \multicolumn{3}{c}{Mean} \\  
  \cmidrule(r){2-4}  \cmidrule(r){5-7}  \cmidrule(r){8-10} 
  & Dice$\uparrow$ & HD95$\downarrow$ & ASSD$\downarrow$ & Dice$\uparrow$ & HD95$\downarrow$ & ASSD$\downarrow$ & Dice$\uparrow$ & HD95$\downarrow$ & ASSD$\downarrow$ \\
  \hline  
  ResUNet  & 96.66 & 6.38 & 1.214
 & 92.40 & \textbf{8.17} & 1.980
 & 94.11 & 7.45 & 1.674 \\
  +SDM
  & \textbf{97.83} & \textbf{3.41} & \textbf{0.760}
 & \textbf{93.47} & 8.37 & \textbf{1.666}
 & \textbf{95.21} & \textbf{6.38} & \textbf{1.303} \\
   \hline 
  nnUNet
 & 97.35 & 4.76 & 0.964
 & 92.96 & 7.41 & 1.562
 & 94.72 & 6.35 & 1.323 \\
 + SDM
 & \textbf{97.42} & \textbf{4.14} & \textbf{0.825}
 & \textbf{93.81} & \textbf{6.98} & \textbf{1.446}
 & \textbf{95.25} & \textbf{5.84} & \textbf{1.198} \\
   \hline 
 MedNeXt
 & 97.42 & 4.57 & 0.907
 & 93.30 & 7.73 & 1.600
 & 94.94 & 6.46 & 1.323 \\
 + SDM
 & \textbf{97.50} & \textbf{4.00} & \textbf{0.671}
 & \textbf{94.13} & \textbf{6.81} & \textbf{1.397}
 & \textbf{95.48} & \textbf{5.68} & \textbf{1.106} \\
   \hline 
 + TransUNet
 & 96.81 & 5.05 & 0.999
 & 92.72 & 8.46 & 1.818
 & 94.35 & 7.10 & 1.490 \\
 + SDM
  & \textbf{97.63} & \textbf{3.57} & \textbf{0.761}
 & \textbf{93.86} & \textbf{7.43} & \textbf{1.437}
 & \textbf{95.37} & \textbf{5.89} & \textbf{1.167} \\
   \hline 
 + UNext
 & 96.03 & 5.67 & 0.949
 & 92.61 & 7.44 & 1.604
 & 93.98 & 6.73 & 1.342 \\
 + SDM
  & \textbf{97.26} & \textbf{5.05} & \textbf{0.824}
 & \textbf{92.67} & \textbf{7.43} & \textbf{1.569}
 & \textbf{94.50} & \textbf{6.48} & \textbf{1.271} \\
   \hline 
  + Swin UNETR
 & 96.21 & 6.41 & 1.089
 & 92.88 & 7.88 & 1.668
 & 94.21 & 7.29 & 1.436 \\
 + SDM
  & \textbf{96.84} & \textbf{5.77} & \textbf{0.965}
 & \textbf{93.27} & \textbf{7.82} & \textbf{1.599}
 & \textbf{94.70} & \textbf{7.00} & \textbf{1.345} \\
  \hline
    \multicolumn{10}{c}{(b) SDM and CCM for fused lung lobes} \\
      \hline  
    ResUNet  & 94.12 & 17.29 & 2.994
 & 55.99 & 39.41 & 8.354
 & 71.24 & 30.56 & 6.210 \\
  + Both
  & \textbf{96.13} & \textbf{6.44} & \textbf{1.137}
 & \textbf{93.45} & \textbf{6.65} & \textbf{1.438}
 & \textbf{94.52} & \textbf{6.56} & \textbf{1.317} \\
   \hline 
  nnUNet
 & 96.51 & 5.77 & 1.081
 & 91.59 & 8.61 & 1.834
 & 93.56 & 7.48 & 1.533 \\
 + Both
 & \textbf{96.74} & \textbf{5.65} & \textbf{0.975}
 & \textbf{92.93} & \textbf{7.32} & \textbf{1.470}
 & \textbf{94.45} & \textbf{6.65} & \textbf{1.272} \\
   \hline 
 MedNeXt
 & 96.78 & 4.92 & 0.968
 & 91.74 & 8.79 & 1.883
 & 93.75 & 7.25 & 1.517 \\
 + Both
 & \textbf{96.86} & \textbf{4.91} & \textbf{0.948}
 & \textbf{92.12} & \textbf{7.78} & \textbf{1.689}
 & \textbf{94.02} & \textbf{6.63} & \textbf{1.393} \\
   \hline 
 + TransUNet
 & 94.82 & 15.80 & 2.339
 & 54.67 & 42.21 & 8.062
 & 70.73 & 31.64 & 5.773 \\
 + Both
  & \textbf{95.55} & \textbf{7.42} & \textbf{1.280}
 & \textbf{84.96} & \textbf{22.64} & \textbf{4.863}
 & \textbf{89.20} & \textbf{16.55} & \textbf{3.430} \\
   \hline 
 + UNext
 & 94.26 & 15.17 & 2.546
 & 85.68 & 25.35 & 4.717
 & 89.11 & 21.28 & 3.848 \\
 + Both
  & \textbf{96.30} & \textbf{5.18} & \textbf{0.992}
 & \textbf{91.31} & \textbf{10.08} & \textbf{2.010}
 & \textbf{93.31} & \textbf{8.12} & \textbf{1.602} \\
   \hline 
  + Swin UNETR
    & \textbf{95.79} & 8.65 & 1.297
 & 88.65 & 11.37 & 2.481
 & 91.51 & 10.28 & 2.007 \\
 + Both
 & 95.76 & \textbf{6.49} & \textbf{1.154}
 & \textbf{90.85} & \textbf{8.64} & \textbf{1.934}
 & \textbf{92.82} & \textbf{7.78} & \textbf{1.622} \\
  \hline  

  \end{tabular}}
  \end{center}
\end{table}


\subsubsection{Ablation on Core Components of the Pulling Branch}
In addition, we make an analysis on the effect of components in CCM. CCM serves as a pulling force to stretch the inter-class boundary region. Besides, each semantic class can be compressed as a more tight object. However, initial class centers face the challenge of slow convergence \citep{meng2021conditional, liu2021dab}, and are significant to the generation of pulling tokens. Thus, we propose the center atlas and pseudo labels for supervision to relieve the convergence difficulty of class centers. Here we carry out a structural ablation study on these two components. As shown in Table \ref{structure ablation}, excluding the center atlas or supervision information will weaken the models' representation ability. Specifically, by removing the loss supervision, there exist $2.11\%\downarrow$ Dice score and $0.62\%\uparrow$ HD95 value for the right middle lobe, $0.64\%\downarrow$ Dice score and $0.20\%\uparrow$ HD95 value for the average metric. And the class center loss can converge well in the training and validation process as referenced in the supplementary material. Besides, after eliminating the center atlas, there are $0.54\%\downarrow$ Dice score and $0.73\%\uparrow$ HD95 value for the right lobe, $0.48\%\downarrow$ Dice score and $0.88\%\uparrow$ HD95 value for the average metric. We provide the T-SNE visualizations of class centers in Figure \ref{class center}. Here sub-figure (a) refers to clustered class centers from the center atlas in the baseline model incorporated with CCM, which represent 6 semantic centers of the lung lobe dataset. And sub-figure (b) is the visualization result of initial class centers in the baseline model. It is shown that clustered class centers can cover a larger area compared with that of initial centers in the baseline model, which indicates a better convergence to the global optimum and a better generalization ability when confronted with various inputs. Besides, the inter-class distance between class centers is larger, which suggests that models are better at addressing boundary confusion.

\subsubsection{Ablation on SDM's plug-and-play Characteristics}
We further probe into the plug-and-play property of SDM on different segmentation networks. Table \ref{tab sdm plug-and-play} lists quantitative comparative results on the clean lung lobe dataset. We can find out that these universal segmentation networks can achieve more precise segmentation predictions when integrated with SDM. In essence, these neural networks adopt a stack of convolutional blocks as the decoder structure, which can aggregate high-level semantics and low-level detail information including shape, texture, color information, etc \citep{ronneberger2015u}. SDM is capable of strengthening raw skipped features from the encoder with abundant boundary and semantic information, and that is why SDM can promote the segmentation performance of baseline models.

\subsubsection{Ablation on plug-and-play Characteristics of the pull-push mechanism}
Similarly, we also investigate the plug-and-play Characteristics of pulling and pushing branches. As shown in Table \ref{tab sdm plug-and-play}, baseline models incorporated with these two branches can achieve better performance on the fused lung lobe dataset, which reveals that the pull-push mechanism bears excellent plug-and-play properties. Specifically, the pull-push mechanism brings significant performance increases for TransUNet and UNeXt, with $18.47\%$ and $4.20\%$ Dice increases respectively. More detailedly, the segmentation task on the right lobes can be better solved, which means that pushing and pulling branches are competent to address boundary confusion. Further, we carry out qualitative analysis on the effectiveness of the pull-push mechanism when plugged into other segmentation models. As shown in Figure \ref{ccm-sdm plug-and-play visualizations}, we select nnUNet \citep{isensee2021nnu}, Swin UNETR \citep{tang2022self}, UNeXt \citep{valanarasu2022unext} from the family of CNNs, Transformer-based and MLP-based models respectively as baseline models. The introduction of SDM and CCM will promote the segmentation precision of inter-class boundary regions, especially in the right lobe.

\section{Conclusion and Future work}
In this paper, we summarize three types of boundary confusion in medical image segmentation and propose PnPNet to address the challenge by modeling the interaction between the boundary and its adjacent regions. Specifically, the pushing and pulling branches are introduced to squeeze and stretch the boundary region. Then these two adversarial forces can boost models' representation abilities on boundaries during training. Our model outperforms other convolutional neural networks (CNNs) and Transformer-based models on four challenging datasets with different types of boundary confusion. Furthermore, extensive experimental results demonstrate that SDM and CCM can boost models' segmentation performance as plug-and-play modules. 

However, due to the fact that the pull-push mechanism is appropriate for addressing the delineation of anatomies with boundary confusion, validation experiments are required on more challenging datasets, even for public datasets from natural scenes. In addition, motivated by Segment-Anything-Model \citep{kirillov2023segment}, delicately designed text prompts will promote models' consistent predictions without outliers, especially for datasets like vertebrae and lung lobes, which need further research.

\section*{Acknowledgments}
This work is supported in part by National Key R\&D Program of China (2019YFB1311503), the Shanghai Sailing Program (20YF1420800), the Shanghai Health and Family Planning Commission (202240110) and Xinhua Hospital affiliated with the School of Medicine (XHKC2021-07).

\bibliographystyle{unsrt}    
\bibliography{main}

\newpage
\appendix

\section*{Supplementary Material}

\section{Structural Details of PnPNet}
In the benchmark of four datasets, we adopt ResUNet \citep{diakogiannis2020resunet} as the baseline model for clean and fused pulmonary lobe datasets, and MedNeXt \citep{roy2023mednext} for VerSe 2019 and LA/LAA datasets. We give a detailed description about the encoder of these two baseline models. For the encoder structure, we provide the scale index, the number of basic blocks in each scale, the type of normalization layers and the number of channels in each block as shown in Table \ref{encoder}. Specifically, the basic block for ResUNet is residual unit \citep{he2016deep}, and the core component for MedNeXt is the MedNeXtBlock. For the decoder structure, we adopt a stack of residual blocks \citep{tang2022self} combined with the upsampling process. For more details of PnPnet, please refer to the source codes released at \url{https://github.com/AlexYouXin/PnPNet}.

\begin{table}[!h]
  \begin{center}
  \caption{The details of encoders in PnPNet, including the scale index, the number of basic blocks in each scale, the type of normalization layers and the number of channels in each block. The parameter number of networks is calculated conditioned on three segmentation classes.}
  \label{encoder}

  \resizebox{0.85\columnwidth}{!}{
  \renewcommand\arraystretch{1.5}
  \begin{tabular}{cccccc}
    \hline    
    Method & Scale & Blocks & Norm Type & Channels & Param (M) \\ 
  \hline  
  ResUNet \citep{diakogiannis2020resunet}  & [$\tfrac{1}{2}$, $\tfrac{1}{4}$, $\tfrac{1}{8}$, $\tfrac{1}{16}$]  & [1, 2, 3, 5] & Batch Norm & [16, 64, 256, 512] & 27.19 \\
  MedNeXt \citep{roy2023mednext} &  [$\tfrac{1}{2}$, $\tfrac{1}{4}$, $\tfrac{1}{8}$, $\tfrac{1}{16}$] & [2, 2, 2, 2] & Group Norm & [32, 64, 128, 256]  & 10.52 \\
  \hline
  \end{tabular}}
  \end{center}
\end{table}

\section{The visualization of center atlas}
\label{sec2}
\begin{figure}[!h]
\centerline{\includegraphics[width=0.60\linewidth]{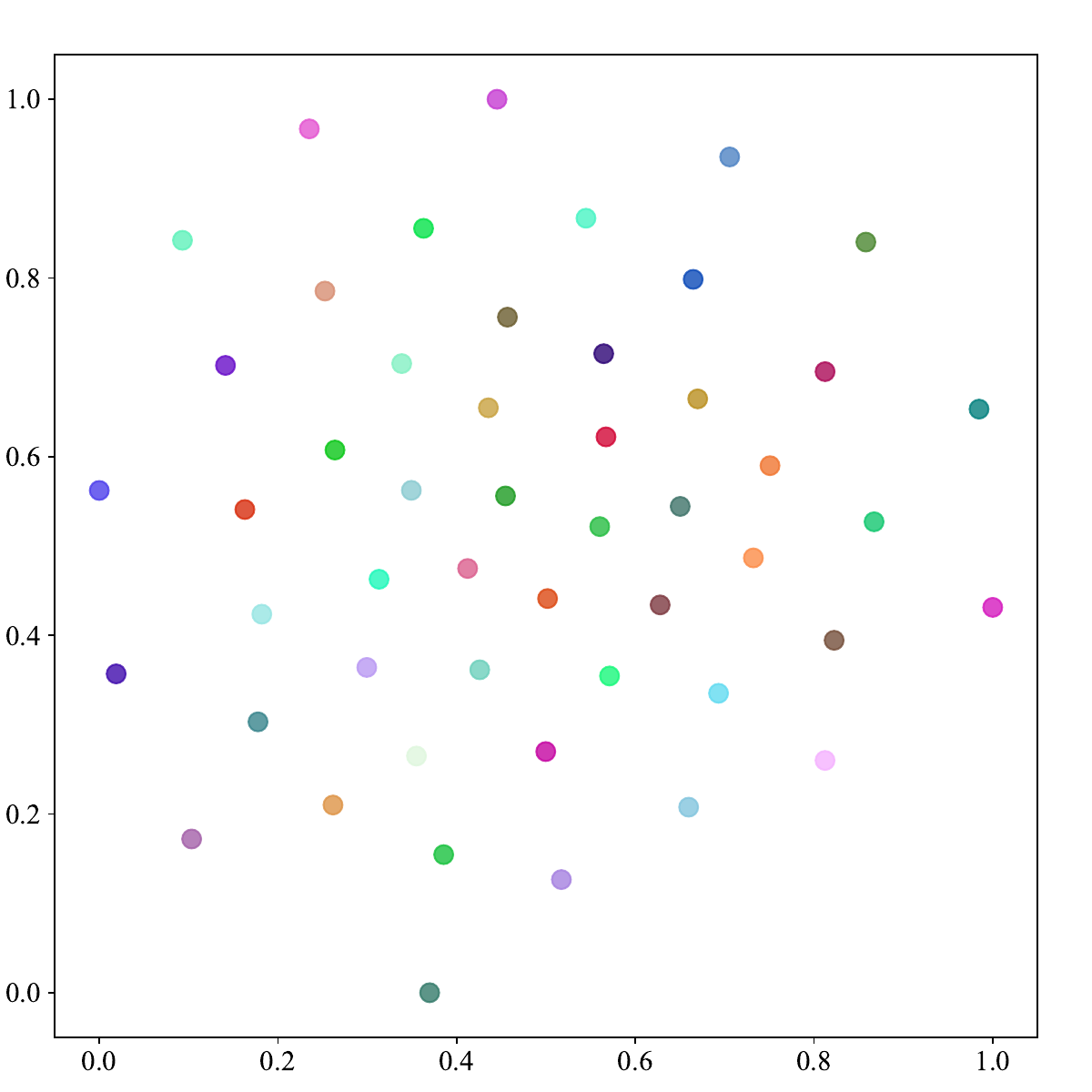}}
\caption{The visualization of center atlas after T-SNE dimension reduction. We can figure out that class centers in the atlas cover almost the whole center space. The distance between adjacent centers is relatively fixed, indicating a small degree of redundancy in deep representations for class centers.}
\label{center atlas}
\end{figure}

Here we probe the distribution of class centers in the center atlas. The number of class centers in the atlas is set as M (M=50), and the dimension of each center is set as 192. The T-SNE process is carried out to present a 2D visualization map. As shown in Figure \ref{center atlas}, class centers in the atlas follow a uniform distribution, in which the distance of adjacent centers is relatively fixed. And the whole distribution region approximates a circle, which means that learnable class centers could generalize well on a large range of dataset distribution statistics.

\section{The visualization of class center loss curves}
In PnPNet, we introduce the class center loss into the pulling branch to accelerate the convergence of learnable class centers \citep{meng2021conditional, liu2021dab}. For the calculation of the class center loss, pseudo labels for class centers are generated by a weighted sum of mask embeddings and skipped features from the encoder. Here this loss exerts semantic prior information into class centers, thus promoting to generate a set of high-quality class centers. As illustrated by Figure \ref{center atlas}, the center loss in training and validation processes shows a fast and stable decrease.

\begin{figure}[!t]
\centerline{\includegraphics[width=0.70\linewidth]{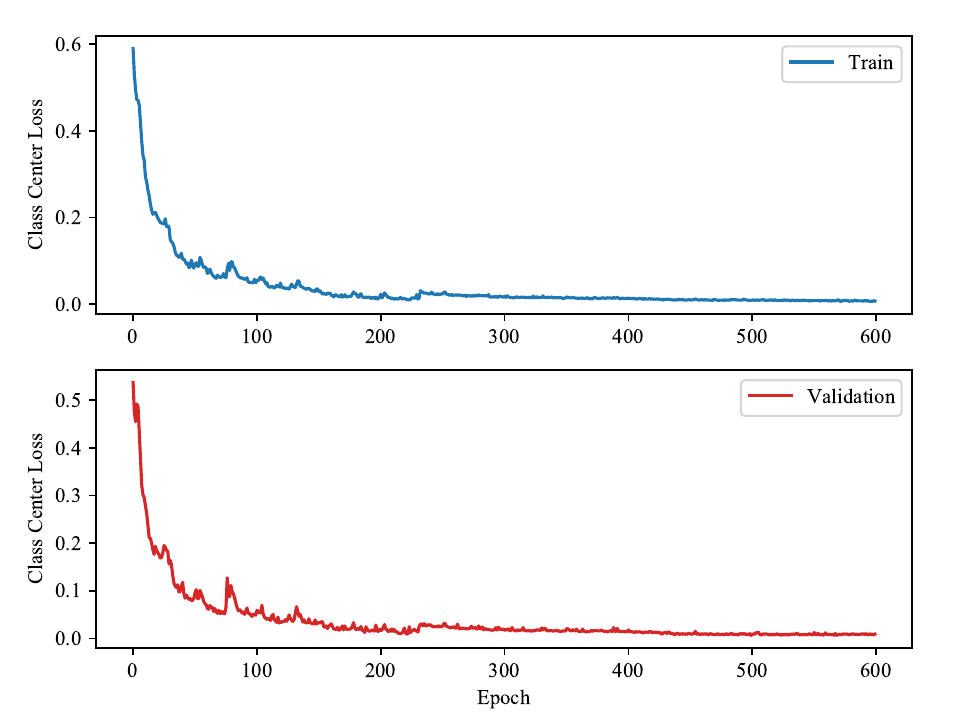}}
\caption{The loss curves for class centers in the training and validation stage.}
\label{center loss}
\end{figure}

\begin{figure}[!t]
\centerline{\includegraphics[width=0.95\linewidth]{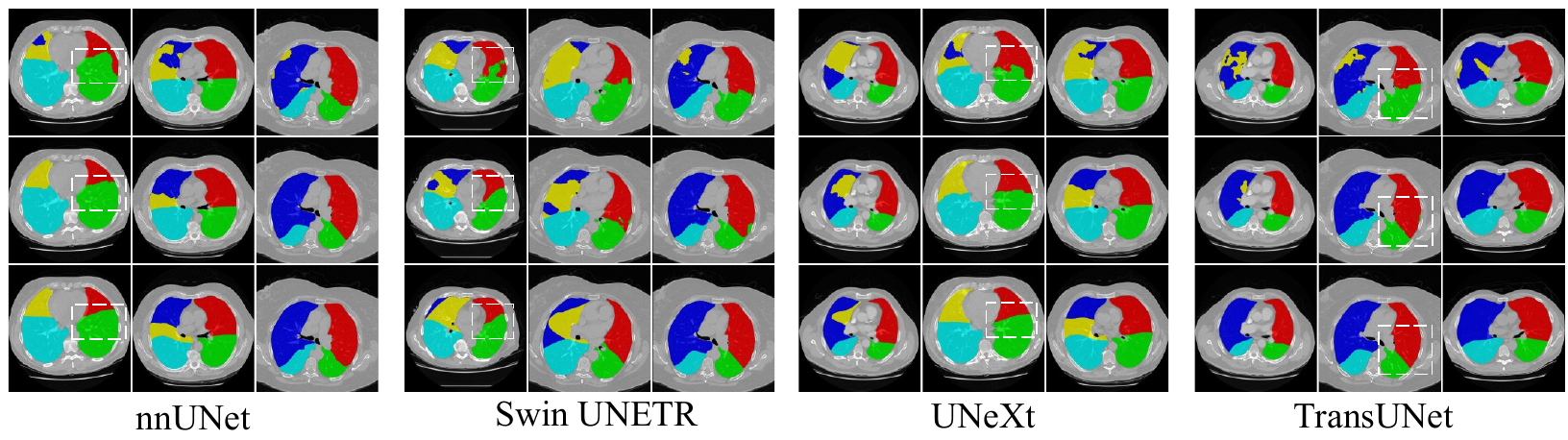}}
\caption{Qualitative visualizations on clean lung lobes when SDM is plugged into other segmentation backbones. Here we choose nnUNet, Swin UNETR, TransUNet, and UNeXt from the family of CNNs, Transformer-based and MLP-based models respectively. The first row: predictions from the baseline model. The second row: predictions from the baseline model plus SDM. The third row: GTs.}
\label{sdm plug-and-play}
\end{figure}

\section{The qualitative analysis for the plug-and-play property of SDM}
\label{sec3}
It has been demonstrated that the semantic difference module bears the plug-and-play property according to quantitative results in Section 4.4.4. SDM shows excellent performance when introduced into ResUNet \citep{diakogiannis2020resunet}, nnUNet \citep{isensee2021nnu}, MedNeXt \citep{roy2023mednext}, TransUNet \citep{chen2021transunet}, Swin UNETR \citep{tang2022self}, and UNeXt \citep{valanarasu2022unext}. In this part, we provide qualitative visualization results on lung lobe predictions. As shown in Figure \ref{sdm plug-and-play}, we figure out that SDM can significantly refine the segmentation masks given by nnUNet \citep{isensee2021nnu}, TransUNet \citep{chen2021transunet}, and UNeXt \citep{valanarasu2022unext}.

\end{document}